\documentclass[preprint,12pt]{elsarticle}

\usepackage{amssymb}
\usepackage{amsmath}

\usepackage{siunitx}
\usepackage{multirow}
\usepackage{tabularx}
\usepackage{booktabs}
\usepackage{array}
\usepackage{makecell}
\usepackage{subcaption}
\usepackage{algorithm}
\usepackage{algorithmic}
\newcolumntype{C}{>{\centering\arraybackslash}X}

\journal{Nuclear Physics B}

\begin{document}

\begin{frontmatter}



\title{A Mechanism and Optimization Study on the Impact of Information Density on User-Generated Content Named Entity Recognition}


\author{Jiang Xiaobo, Dinghong Lai, Song Qiu, Yadong Deng, Xinkai Zhan} 

\affiliation{organization={},
            addressline={}, 
            city={},
            postcode={}, 
            state={},
            country={}}

\begin{abstract}
Named Entity Recognition (NER) models trained on clean, high-resource corpora exhibit catastrophic performance collapse when deployed on noisy, sparse User-Generated Content (UGC), such as social media. Prior research has predominantly focused on point-wise symptom remediation---employing customized fine-tuning to address issues like neologisms, alias drift, non-standard orthography, long-tail entities, and class imbalance. However, these improvements often fail to generalize because they overlook the structural sparsity inherent in UGC. This study reveals that surface-level noise symptoms share a unified root cause: low Information Density (ID). Through hierarchical confounding-controlled resampling experiments (specifically controlling for entity rarity and annotation consistency), this paper identifies ID as an independent key factor. We introduce Attention Spectrum Analysis (ASA) to quantify how reduced ID causally leads to ``attention blunting,'' ultimately degrading NER performance. Informed by these mechanistic insights, we propose the Window-Aware Optimization Module (WOM), an LLM-empowered, model-agnostic framework. WOM identifies information-sparse regions and utilizes selective back-translation to directionally enhance semantic density without altering model architecture. Deployed atop mainstream architectures on standard UGC datasets (WNUT2017, Twitter-NER, WNUT2016), WOM yields up to 4.5\% absolute F1 improvement, demonstrating robustness and achieving new state-of-the-art (SOTA) results on WNUT2017.

\end{abstract}





\begin{keyword}

Noisy User-Generated Content, Named Entity Recognition, Mechanistic Analysis, Information Density, Window-aware Optimization

\end{keyword}

\end{frontmatter}



\section{Introduction}
\label{sec1}

Named Entity Recognition (NER), a core task in Natural Language Processing (NLP), provides critical support for information extraction, knowledge graph construction, and downstream analysis\cite{seow2025review}. With the explosive growth of User-Generated Content (UGC) such as social media and online reviews, efficiently extracting valuable entity information from massive noisy text has become a focal point for both academia and industry. UGC introduces new challenges: open-world new entities and alias drift, timeliness and cross-domain issues, non-orthography and tokenization fragility, and long-tail entities/class imbalance. These new challenges cause a sharp decline in NER performance\cite{ushio2022named}.

To address these challenges, new algorithms have been proposed\cite{keraghel2024survey}. For non-standard text, models such as TweetBERT are pre-trained on social media corpora to better capture informal language patterns\cite{qudar2020tweetbert}, while joint character–word encoding improves robustness to orthographic variation\cite{Li2020FLAT}. For open-world entities, methods include entity-aware representations\cite{lu2022unified} and retrieval-augmented generation with knowledge bases\cite{brisson2025named, mahtab2025bannerd}. Additional strategies include multi-granularity feature fusion\cite{Li2023Multigranularity}, domain adversarial training for cross-domain sparsity\cite{Chen2021AdvPicker}, and prompt-based few-shot adaptation\cite{wei2022chain}.

However, despite continuous improvements in model architecture, performance gains on UGC have hit a bottleneck. The main reasons are: (1) Lack of in-depth analysis of UGC's impact on performance; even if an improvement is proposed for one factor, performance easily degrades when encountering others. (2) Lack of mechanistic analysis of the performance impact; the proposed countermeasures sometimes fail to solve the problem.

Looking beyond these surface-level factors, this paper identifies the core issue as the deeper, structurally-related concept of “information density” (ID). This paper defines ID as the strength with which an entity signal is supported by effective contextual cues in its local textual environment. Unlike the traditional simple entity ratio, ID calculation focuses on the local environment around the entity and weights contextual importance, thus more accurately measuring the clarity of the entity signal. For instance, consider the formal text \textit{Do you have tickets for Madison Square Garden?}" versus the UGC snippet \textit{Got tix 4 msg?}". In the latter case, the entity "msg" is surrounded by low-information particles without sufficient syntactic anchors. Unlike general OCR noise which represents character-level corruption, this reflects a structural scarcity of semantic support.

To systematically examine the role of information density, this paper isolates ID as an independent structural factor through confounding-controlled resampling, ensuring that performance variations are not driven by entity rarity or annotation inconsistency. Building on this analysis, we propose a model-agnostic Window-aware Optimization Module (WOM) that enhances effective information density by suppressing uncontrollable noise and semantic perturbations during encoding.The main contributions of this work are summarized as follows:

(1)Definition and Importance Analysis of Information Density: This paper defines and quantifies information density. Through correlation analysis and global sensitivity analysis (based on Morris and Sobol methods), it preliminarily verifies that information density is a key structural factor influencing model performance.

(2)Mechanistic Dissection and Deep Revelation: This paper goes beyond surface-level correlation to deeply dissect two core pathways through which low information density damages model performance. We innovatively propose the Attention Spectrum Analysis (ASA) metric, revealing that a statistical conservative bias dominated by background tokens systematically suppresses recall during training. We also quantify and confirm from a frequency-domain perspective that low information density causes “attention blunting", weakening the model's ability to focus on local key information.

(3)Targeted Optimization Strategy: This paper propose WOM, an LLM-empowered, model-agnostic framework that selectively repairs information-sparse contextual windows at the data level. By dynamically identifying entity-centered windows and amplifying effective contextual signals while suppressing irrelevant noise, WOM precisely enhances effective information density without modifying the underlying model architecture. Experiments on multiple benchmarks, including WNUT2017\cite{derczynski2017}, show consistent improvements of 1.0–4.5\% F1 across architectures, achieving a new state of the art on WNUT2017.

The remainder of this paper is organized as follows. Section II reviews related work on noisy-text NER and dataset structural analysis. Section III defines the proposed features and evaluation metrics. Section IV presents correlation and sensitivity analyses, along with a mechanistic study of how information density influences attention and decision-making. Section V introduces the WOM model. Section VI reports experimental results across multiple architectures, including ablation and hyperparameter analyses. Section VII concludes the paper.



\section{Related Work}
\label{sec2}
\subsection{NER Model Evolution and Limitations}
Transformer-based Pre-trained Language Models (PLMs) have propelled the NER field into a new paradigm\cite{vaswani2017}. Large-scale PLMs represented by BERT\cite{devlin2019} and RoBERTa\cite{liu2019} have significantly improved entity recognition accuracy through large-scale unsupervised pre-training and downstream NER task fine-tuning, repeatedly setting SOTA records on standard benchmarks like CoNLL-2003 and OntoNotes. Subsequent research continues to innovate: LUKE\cite{yamada2020luke} explicitly introduced entity-aware self-attention, and DeBERTa\cite{he2021deberta} further enhanced semantic understanding for NER by improving the attention mechanism and pre-training paradigm.

Currently, the generalization ability of PLMs has been validated in scenarios like multilingual and multi-domain NER, e.g., BioBERT's excellent performance on biomedical literature\cite{lee2020biobert}, proving PLMs' high accuracy in specialized, standardized texts with regular terminology. However, in noisy texts with sparse entity signals, such as social media, these methods still face performance bottlenecks. Relying solely on model architectures that perform well on standard data cannot guarantee success in challenging new scenarios. In other words, performance issues stem not only from model architecture limitations but also from the imbalanced information structure within the input data. Therefore, this study does not solely seek breakthroughs in model structure; rather, it systematically reveals the deep mechanistic impact of information density on model performance and aims to specifically enhance the model's NER capabilities in low-density regions.

\subsection{Data-Driven Optimization Strategies on NER}
For noisy texts like social media has been an important agenda in the NLP field for the last decade. Texts from social media platforms like Twitter and Reddit are often characterized by informal expression and diverse spelling formats, significantly increasing recognition difficulty. Although pre-trained models have become the mainstream solution for NER, they still struggle to adapt to common text noise due to tokenization limitations\cite{srivastava2020noisy}. Some studies have begun to focus on various data-level enhancement strategies. To address novel and rare entities, researchers have made improvements by introducing contextual modeling\cite{esmaail2024ner} or retrieval augmentation\cite{monajatipoor2024llms}. Facing the generalization dilemma in low-resource texts, transfer learning has also been widely adopted\cite{akkaya2021transfer},\cite{zhang2023extractive},\cite{kirsch2023towards}. These techniques provide effective solutions for improving UGC performance, but the lack of a decision-making basis makes it difficult to select a suitable strategy based on data characteristics.

\subsection{Diagnostic Analysis of Performance Impact}
The concept of data-centric AI is reshaping the AI research landscape~\cite{whang2023data}, \cite{zha2025data-centric}. Unlike model-driven NER, this new paradigm emphasizes diagnosing data properties to uncover performance bottlenecks. Existing research has explored the impact of specific data factors: Zhu et al.\cite{zhu2023investigating}  systematically analyzed the impact of annotation noise on NER model performance, proposing targeted label denoising and sample re-weighting strategies; Eisape et al.\cite{eisape2022probing} proposed “probing" to detect whether models internally encode entity knowledge and to causally intervene in model representations. These works all attempt to establish a link from quantifiable data properties to internal model behaviors\cite{xiaobo2025relation},\cite{lai2026mechanistic}. However, they have two limitations: In terms of breadth, the analyzed features are often single variables, lacking a systematic framework to comprehensively evaluate and compare the combined impact of various features. In terms of depth, most stop at the observed phenomena, still lacking exploration into the influence mechanisms of text structural features.

\section{Preliminaries}
\label{sec3}
This section elaborates on the structural features and evaluation metrics. To move beyond phenomenological descriptions and reveal the fundamental data structural factors affecting model performance, this paper defines and quantifies six structural feature metrics to measure the intrinsic complexity of a dataset. Concurrently, the entire paper employs the F1-score as the core performance evaluation metric, aiming to systematically analyze the correlation between these structural features and model generalization performance.

\subsection{Structural Feature Definitions}
The paper defines and quantifies 6 structural features for subsequent experimental calculation and validation.

\subsubsection{Information Density (NED)}
Measures the concentration of entity information. Let $ET$ be the number of entity tokens and $TT$ be the total token count. While $\frac{ET}{TT}$ is intuitive, it ignores sentence length differences. Thus, this paper introduces a sentence-length-based correction term. The definition of Information Density is:

\begin{equation}
\label{deqn_ex1}
\text{NED} = \frac{ET}{TT} \Bigl(1+\log(TT_{SL})\cdot{\lambda}\Bigl)
\end{equation}

where $TT_{SL}$ is the total token count in sentences containing entities, and $\lambda$ is the text structure factor (default 0.1).

\subsubsection{Entity Imbalance Degree}
Normalized standard deviation (NormSTD) is a statistical measure used to assess the extent of imbalance in the distribution of different entity categories.
For a proportion vector $\mathbf{p} = (p_1, ..., p_C)$ of C entity categories, where $p_i = \frac{n_i}{\sum_j n_j}$, NormSTD is defined as:

\begin{equation}
sigma = \sqrt{\frac{1}{C}\sum_{i=1}^C \left(p_i - \frac{1}{C}\right)^2} \label{2a} 
\end{equation}

\begin{equation}
\text{NormSTD} = \frac{\sigma}{\sigma_{\max}} = \sigma \times \frac{C}{\sqrt{C-1}} 
\end{equation}

\subsubsection{Redundancy}
This index measures the proportion of duplicate samples in the dataset $D$.
A lower redundancy value indicates a reduced number of duplicates and increased data diversity.
It is defined as:

\begin{equation}
\text{Red}(D) = 1 - \frac{|\{(x^{(i)}, y^{(i)}) : i=1, ..., n\}|}{n}
\end{equation}

\subsubsection{Entity Polysemy}
The Normalized Label Entropy per Entity (ELE) metric is employed to quantify this phenomenon.
The flatter the label distribution, the more severe the polysemy.
For each entity $e$, let $\{n_{e,1}, n_{e,2}, ..., n_{e,C}\}$ be the number of times it is annotated with different categories in the corpus, and $N_e = \sum_i n_{e,i}$ be its total occurrences.
The label distribution entropy for this entity is as follows:

\begin{equation}
H(e) = -\sum_{i=1}^c \frac{n_{e,i}}{N_e} \log\left(\frac{n_{e,i}}{N_e}\right)
\end{equation}

\begin{equation}
\text{ELE} = \frac{1}{E} \sum_{e=1}^E \tilde{H}(e) = \frac{1}{E} \sum_{e=1}^E \frac{H(e)}{\log C}
\end{equation}

\subsubsection{Subword Segmentation Rate}
Subword Segmentation Rate reflects the novelty of entity contexts.
A higher value indicates that words are, on average, segmented into more subwords by the tokenizer, suggesting they are rarer.
We simulate the tokenization results of a pre-trained language model on the dataset's tokens using a method that approximates WordPiece segmentation.
It is defined as:

\begin{equation}
\text{SSR} = \frac{\sum_{i=1}^{T} k_i}{T}
\end{equation}

\subsubsection{Lexical Shannon Entropy}
This is the information entropy calculated from the distribution of all tokens, used to measure the lexical diversity of the entire dataset.
It is defined as:

\begin{equation}
\text{VocabEntropy} = H_{\text{token}} = -\sum_{i=1}^{n} p_i \log_2 p_i
\end{equation}

\subsection{Datasets}
To systematically study the characteristics of noisy User-Generated Content (UGC) and validate the effectiveness of the proposed method, this study selects three publicly recognized benchmark datasets in this domain. These datasets all originate from social media platforms and share common challenges such as short texts, informal expressions, and sparse entity signals, making them ideal choices for testing NER model robustness.

WNUT2017 is the core dataset for our main analysis and experiments, and one of the most challenging benchmarks in the noisy text NER field. This dataset contains English tweets from Twitter, covering emerging and rare entity types. Its noise characteristics and low-information-density environment pose a severe test for NER models. WNUT2016, as the previous version of the WNUT series, also consists of Twitter tweets and is highly similar to WNUT2017 in data distribution and challenges. We use it as one of the validation datasets to test the universality of our proposed analysis framework and optimization module. Twitter-NER  is another dataset widely used for UGC NER tasks. Similar to the WNUT series, it also contains a large amount of unstructured social media text. We use this dataset to further verify the generalization ability and effectiveness of the WOM module under different data distributions.

\subsection{Model Methods}
To ensure that the conclusions of our diagnostic framework and optimization strategy are generalizable across architectures, we selected two mainstream NER models representing different technical paradigms as our core analysis tools, and extended to multiple baseline models for performance evaluation in the experimental section.

RoBERTa-BiLSTM-CRF\cite{RoBERTa-BiLSTM-CRF2024xu}: This model is a classic and powerful sequence labeling architecture. It first utilizes a Robustly Optimized BERT Pretraining Approach (RoBERTa) as an encoder to transform input text into deep, context-rich semantic representations. RoBERTa, by optimizing BERT's pre-training strategy (e.g., dynamic masking, NSP task removing), has shown stronger performance on many NLP tasks. Subsequently, a Bidirectional Long Short-Term Memory (BiLSTM) layer is used to capture long-range dependencies in the token sequence output by RoBERTa, integrating sequence information from both forward and backward directions. Finally, a Conditional Random Field (CRF) layer is placed on top to learn constraints between labels (e.g., “B-PER" cannot be directly followed by “I-LOC"), thus ensuring the output entity label sequence is structurally valid. The Roberta-BiLSTM model mentioned in this paper is a simplified version of this architecture.

SpanNER\cite{fu2021spanner}: This model represents another mainstream NER paradigm, which models the entity recognition task as a span prediction problem. Unlike traditional sequence labeling models that tag tokens one by one, SpanNER directly extracts entities by predicting their start and end boundaries in the text. This model first uses a pre-trained model like BERT to obtain contextual representations, then designs a module to enumerate all possible spans and calculate the probability of each span being a specific entity type. This method naturally avoids the problem of invalid label sequences and has potential in complex scenarios like recognizing discontinuous or nested entities.

In addition to the two core analysis models above, in the experimental evaluation in Section VI, we also apply WOM to various baseline models including BERT, MINER, etc., to comprehensively validate its performance gains.

\subsection{Evaluation Metrics}
The F1 score is utilized as the primary metric for performance evaluation. It is the harmonic mean of precision and recall, providing a comprehensive measure of performance.

\begin{equation}
precision = \frac{TP}{TP + FP}
\end{equation}

\begin{equation}
recall = \frac{TP}{TP + FN}
\end{equation}

\begin{equation}
F1 = 2 \times \frac{precision \times recall}{precision + recall}
\end{equation}

TP (true positives), FP (false positives), and FN (false negatives) represent the match between predictions and ground-truth annotations.

\section{Mechanistic Analysis}
Noisy User-Generated Content (UGC) faces many challenges, such as non-orthography and long-tail entities, but its structural deficiencies are not yet well-understood. Preliminary observations of datasets like WNUT2017 reveal a common phenomenon: significant model performance decline is more often concentrated in simple sentences where entity signals are diluted by a large number of non-entity tokens. Therefore, this paper proposes a core hypothesis: Information Density (ID), the concentration of entity signals in their local context, is the dominant factor constraining NER performance. To systematically validate this hypothesis and reveal its internal mechanisms, this chapter will conduct a progressive, three-level analysis using two representative NER models—BERT-BiLSTM and SpanNER. These two models were chosen because they represent the mainstream sequence labeling and span classification approaches, respectively; their complementary methods help test the robustness of our diagnostic framework across different paradigms. The specific analysis is as follows: This section conducts a mechanistic analysis at three progressive levels. First, through correlation analysis, we preliminarily probe the statistical association between structural features and performance metrics. Next, through sensitivity analysis, we deeply evaluate the model performance's responsiveness to changes in different structural features. Finally, through impact pathway analysis, we reveal the intrinsic pathways through which key structural features affect NER performance. This chapter will detail the analysis using the WNUT2017 dataset (results for other datasets are in the Appendix), aiming to profoundly reveal the structural roots of NER performance issues in UGC data through this analytical chain, moving from surface to depth, from correlation to causation.

\subsection{Correlation Analysis}

\begin{figure}[!htbp]
\centering
\includegraphics[width=\linewidth]{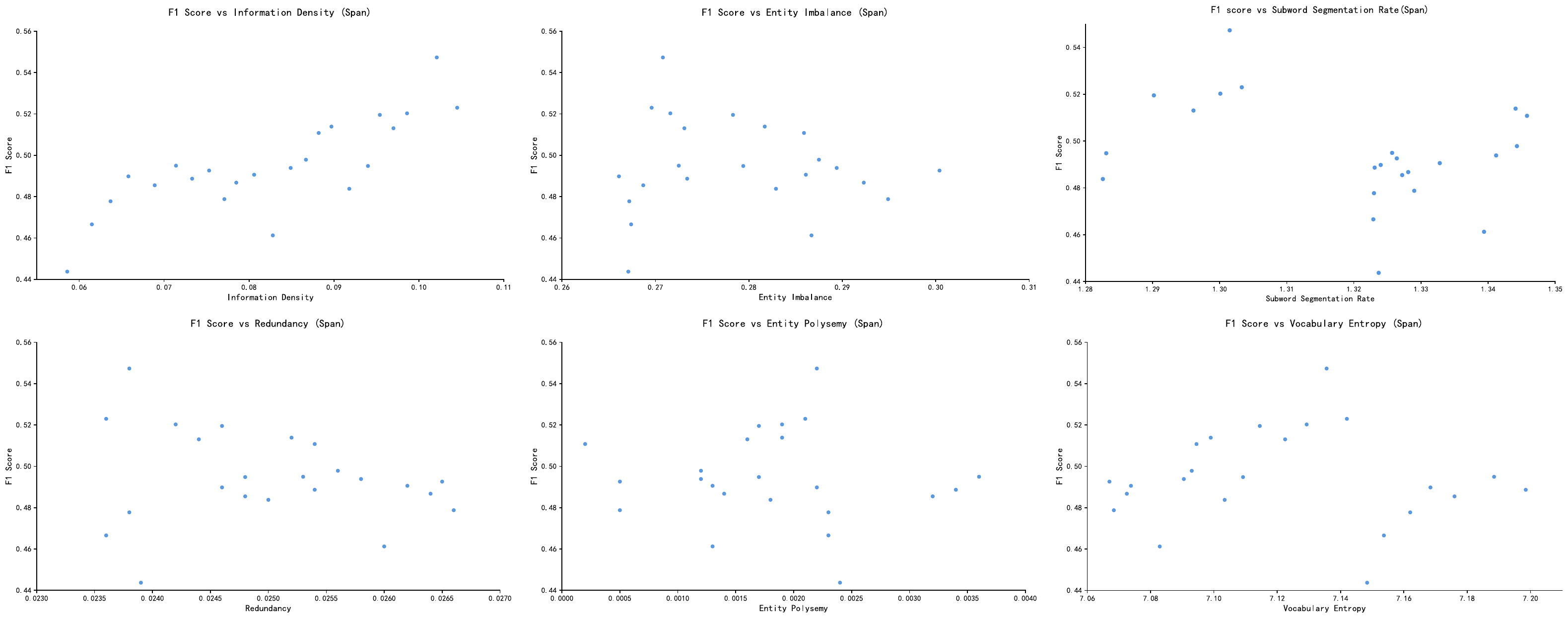}
\caption{Figure Caption}\label{span}
\end{figure}

\begin{figure}[!htbp]
\centering
\includegraphics[width=\linewidth]{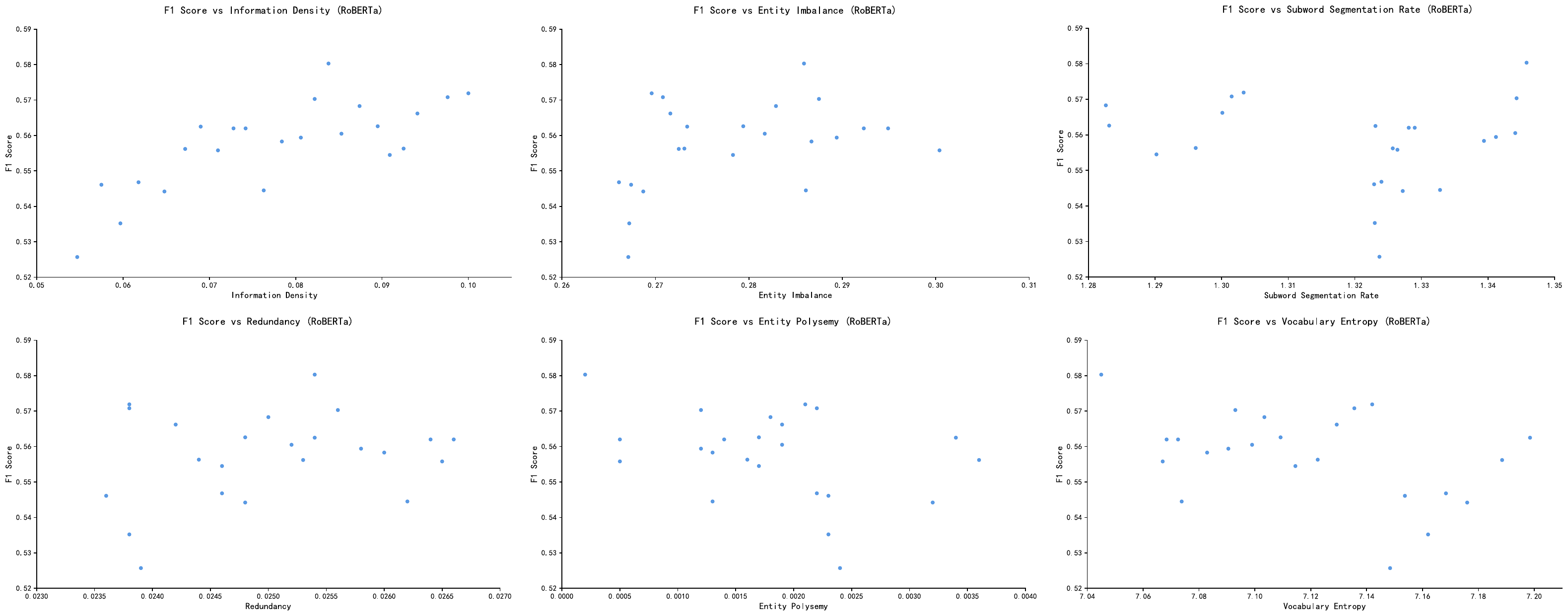}
\caption{Figure Caption}\label{lstm}
\end{figure}

This section first constructs 23 subsets with diverse distributions of structural features, such as information density, entity imbalance and redundancy by performing stratified random sampling on the original training set. Subsequently, two representative models, Roberta-BiLSTM and SpanNER, are independently trained and evaluated on each subset. The relationship between structural features and F1-scores is visualized using scatter plots.

\begin{table}[!t]
\centering
\small
\sisetup{table-format = -1.3, group-digits=false}
\setlength{\tabcolsep}{4pt} 
\begin{tabularx}{\linewidth}{>{\centering\arraybackslash}p{3.8cm} C S[table-format=-1.3] S[table-format=-1.3]}
    \hline
    \textbf{Structural Feature} & \textbf{Method} & \multicolumn{1}{c}{\makecell[c]{\textbf{SpanNER}\\\textbf{(bert-large)}}} & \multicolumn{1}{c}{\makecell[c]{\textbf{Roberta(base)}\\\textbf{BiLSTM}}} \\
    \hline
    \multirow{2}{=}{\centering Information Density} & Pearson & 0.797 & 0.774 \\
    & Sperman & 0.791 & 0.721 \\
    \hline
    \multirow{2}{=}{\centering Entity Imbalance Degree} & Pearson & -0.048 & 0.363 \\
    & Sperman & 0.025 & 0.322 \\
    \hline
    \multirow{2}{=}{\centering Subword Segmentation Rate} & Pearson & -0.323 & -0.216 \\
    & Sperman & -0.147 & -0.133 \\
    \hline
    \multirow{2}{=}{\centering Redundancy} & Pearson & -0.187 & 0.228 \\
    & Sperman & -0.184 & 0.086 \\
    \hline
    \multirow{2}{=}{\centering Entity Polysemy} & Pearson & -0.004 & -0.238 \\
    & Sperman & -0.072 & -0.235 \\
    \hline
    \multirow{2}{=}{\centering Lexical Shannon Entropy} & Pearson & -0.034 & -0.303 \\
    & Sperman & -0.005 & -0.211 \\
    \hline
\end{tabularx}
\caption{Correlation Analysis}\label{table1}
\end{table}

To precisely quantify this degree of association, this paper calculates the Pearson and Spearman correlation coefficients. This method aims to reveal the statistical dependencies between dataset structural features and model performance, providing a rigorous quantitative tool for diagnosing how data properties constrain model performance.

The experimental results show that among all structural features, information density has the strongest correlation with model performance. The scatter plots in Fig.\ref{span} and Fig.\ref{lstm} clearly reveal a significant positive correlation between information density and F1-score for both Roberta-BiLSTM and SpanNER models. Quantitative analysis in Table \ref{table1} further validates this observation: for the Roberta-BiLSTM model, the Pearson and Spearman correlation coefficients reached 0.774 and 0.721, respectively; for the SpanNER model, the correlation was even more pronounced, with coefficients as high as 0.797 and 0.791. The p-values for both tests were well below 0.01, indicating that this positive correlation is highly statistically significant. This strong empirical result not only confirms our core hypothesis—that the dataset's intrinsic structure is a fundamental reason constraining model performance—but also highlights the unique role of information density as a key driving factor.

\subsection{Sensitivity Analysis}

\begin{figure}[!htbp]
\centering
\subfloat[]{\includegraphics[width=0.48\linewidth]{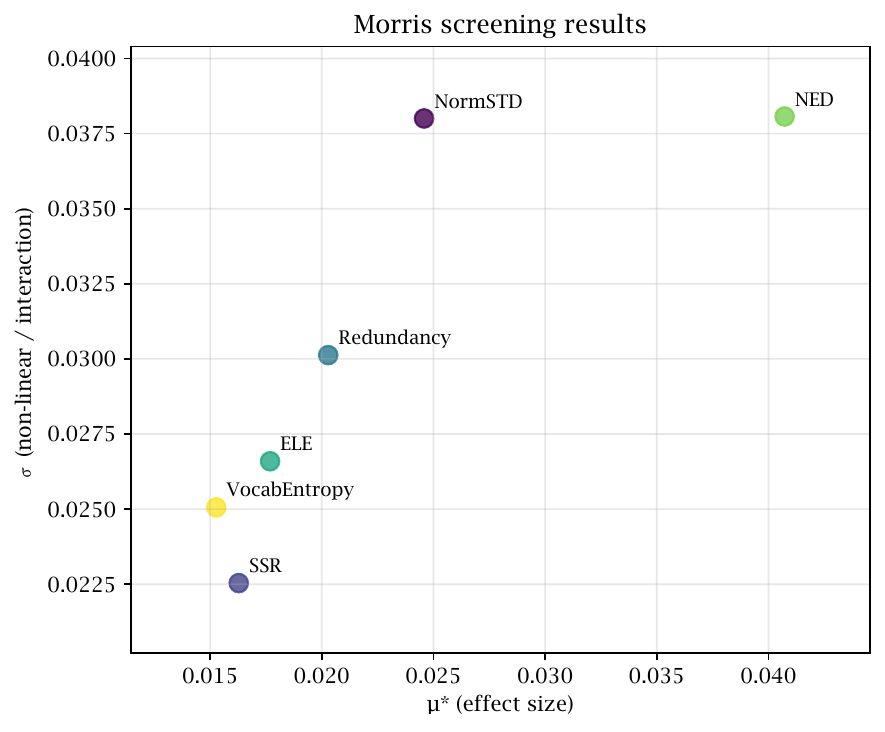}}%
\label{fig_first_case1}
\hfil
\subfloat[]{\includegraphics[width=0.48\linewidth]{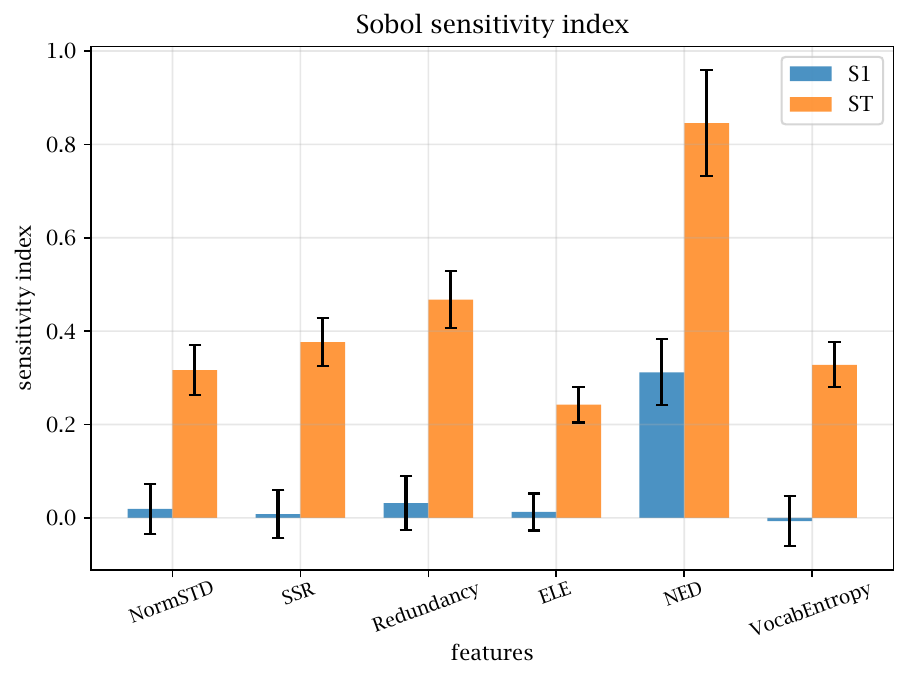}}%
\label{fig_second_case1}
\caption{SpanNER: Sensitivity analysis results for Morris and Sobol using different structural features. (a) Morris. (b) Sobol.}
\label{sen1}
\end{figure}

\begin{figure}[!htbp]
\centering
\subfloat[]{\includegraphics[width=0.48\linewidth]{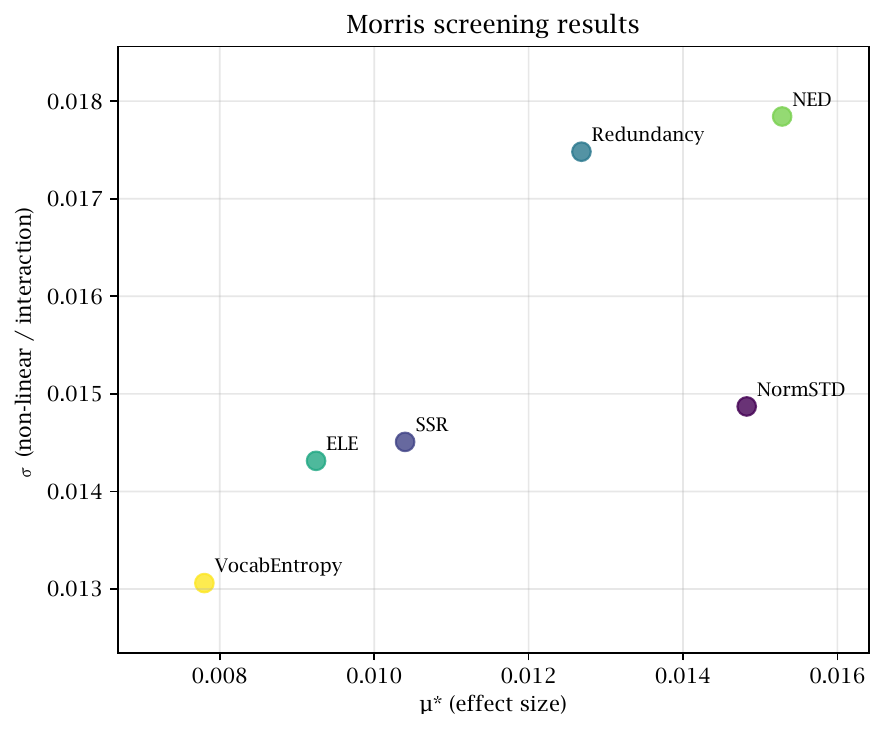}}%
\label{fig_first_case2}
\hfil
\subfloat[]{\includegraphics[width=0.48\linewidth]{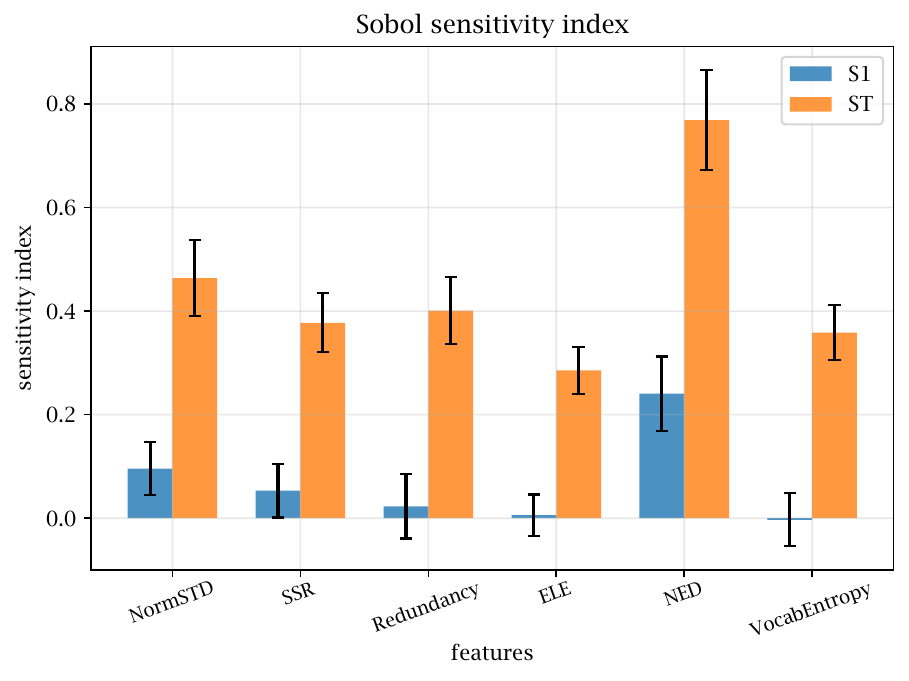}}%
\label{fig_second_case2}
\caption{RoBERTa-BiLSTM: Sensitivity analysis results for Morris and Sobol using different structural features. (a) Morris. (b) Sobol.}
\label{sen2}
\end{figure}

To further verify that information density is the structural feature with the most significant impact on model performance, the experiment employs Global Sensitivity Analysis (GSA). Specifically, this section combines the Morris screening method and the Sobol variance decomposition method. The Morris method uses a method based on probability uniform sampling to quantify the contribution of model input factors to the output. Its principle is close to the one-factor-at-a-time (OAT) method in local sensitivity analysis but captures non-linear and interaction effects through multiple random trajectory samplings. Sobol analysis, based on variance decomposition, further quantifies the main effects and total effects of features to evaluate their independent and interactive contributions, providing a rigorous framework.

The sensitivity analysis is based on the data and models employed in the Section IV.A correlation analysis, implemented using Python's SALib module. Additional results are detailed in the appendix. Sobol indices are estimated using Saltelli's quasi-Monte Carlo extension based on Sobol low-discrepancy sequences; Morris screening employs the classic OAT design, generating random trajectories on a hierarchical grid with $p=6$.

The GSA results are shown in the figures. Fig.\ref{sen1} and Fig.\ref{sen2} show the Morris and Sobol analysis results under the SpanNER and RoBERTa-BiLSTM models, respectively. The $\mu^*-\sigma$ plots on the left show the overall impact (x-axis) and the non-linear or interaction strength (y-axis) of different structural features on model performance. On both models, the mean and standard deviation of the base effect for information density (labeled NED in the figure) are particularly prominent among all features. The plots on the right visualize the Sobol S1 (first-order index) and ST (total-order index) (Note: error bars are 95\% CI). The analysis results are highly consistent with the Morris method and provide deeper insights. Among all structural features, only information density (NED) shows a significant first-order effect in both models; other factors contribute mainly through interactions (ST $\gg$ S1). These experimental results indicate that information density is not only the primary factor affecting the model's F1-score, but its influence is also highly non-linear, with complex interactions with other structural features (e.g., entity imbalance, redundancy).

Based on the results of these two methods, this experiment draws a cross-model, highly consistent conclusion: Information density is the most critical structural factor constraining model performance in UGC scenarios. However, GSA is essentially still attribution analysis, only quantifying the strength of the input-output correlation, without revealing the internal pathways. To fundamentally understand and intervene in performance bottlenecks, it is still necessary to delve into the model's internals to analyze the mechanism by which low information density affects the model's learning process.

\subsection{Impact Mechanism Analysis}
Based on the quantitative analysis in the previous section, this study focuses on the qualitative and mechanistic analysis of the impact of Information Density (ID), aiming to deeply dissect the pathways through which low ID negatively affects model performance, providing solid theoretical support for our findings. To investigate the impact of different ID distributions, this section designs the following strategy using stratified random sampling on the original WNUT2017 training set: We divide the original dataset $D$ into an entity-containing subset $D_E$ and an entity-free subset $D_O$. By applying different sampling rates $p$ ($p \in [0.5, 1.0]$) to randomly downsample $D_E$ to get $D_E'$, and then merging with the complete $D_O$, i.e., $D_p = D_E' \cup D_O$, we generate a series of datasets $\{D_p\}$ with controlled entity ratios. This section will explore two key pathways: first, statistical bias and the loss function, and second, the frequency-domain response of the self-attention mechanism; we will detail how low ID affects the loss function during model optimization and the operation of the Transformer's internal self-attention mechanism.

\subsubsection{Statistics and Loss: Performance Erosion by Background Tokens}
\begin{figure}[!htbp]
\centering
\includegraphics[width=\linewidth]{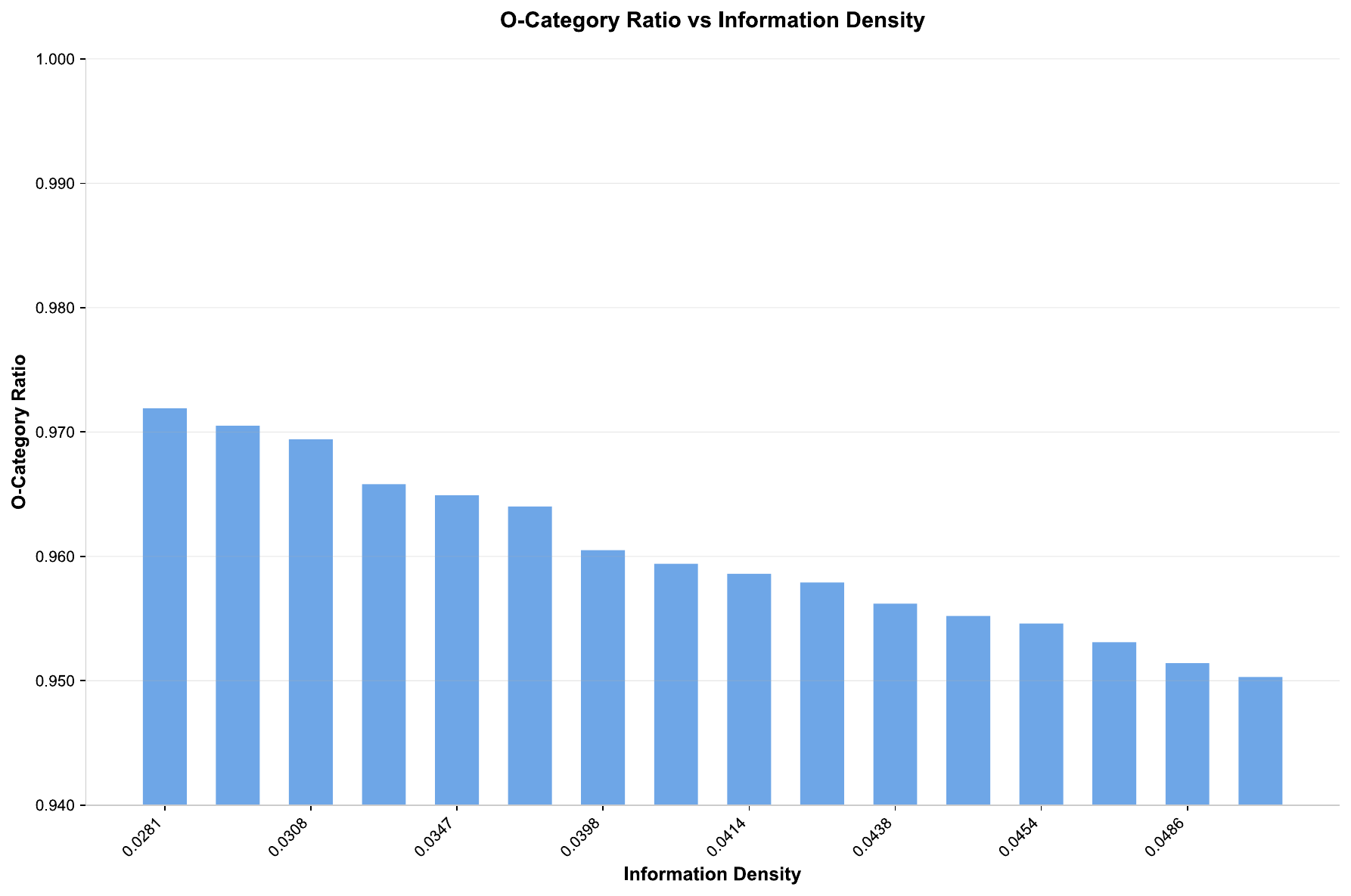}
\caption{Relationship between Information Density and Proportion of O Category.}
\label{Olei}
\end{figure}

Low ID causes an extreme imbalance between entity and background tokens at the data level, inducing severe statistical bias, which is an endogenous root cause of model performance degradation. In low-ID datasets, the non-entity ('O') label holds an overwhelming majority, forming a typical long-tail distribution. Our statistical analysis of the relationship between information density and 'O' proportion in subsets $\{D_p\}$ finds that as information density decreases, the proportion of non-entity labels gradually increases (as shown in Fig.\ref{Olei}, WNUT2017 dataset statistics), meaning the vast majority of samples processed during model training are background tokens.

\begin{figure}[!htbp]
\centering
\includegraphics[width=0.48\linewidth]{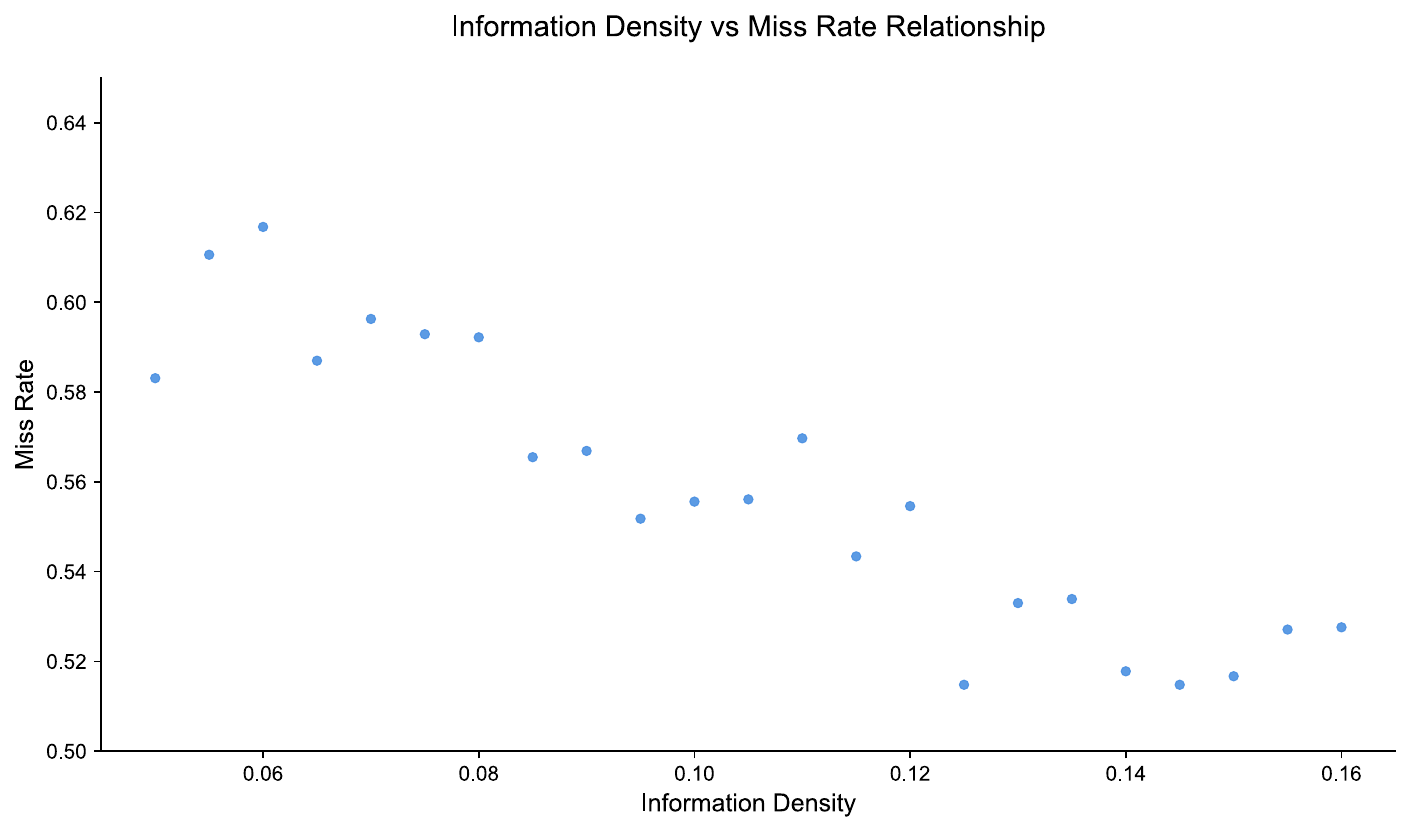}
\label{inforfig1}
\hfil
\includegraphics[width=0.48\linewidth]{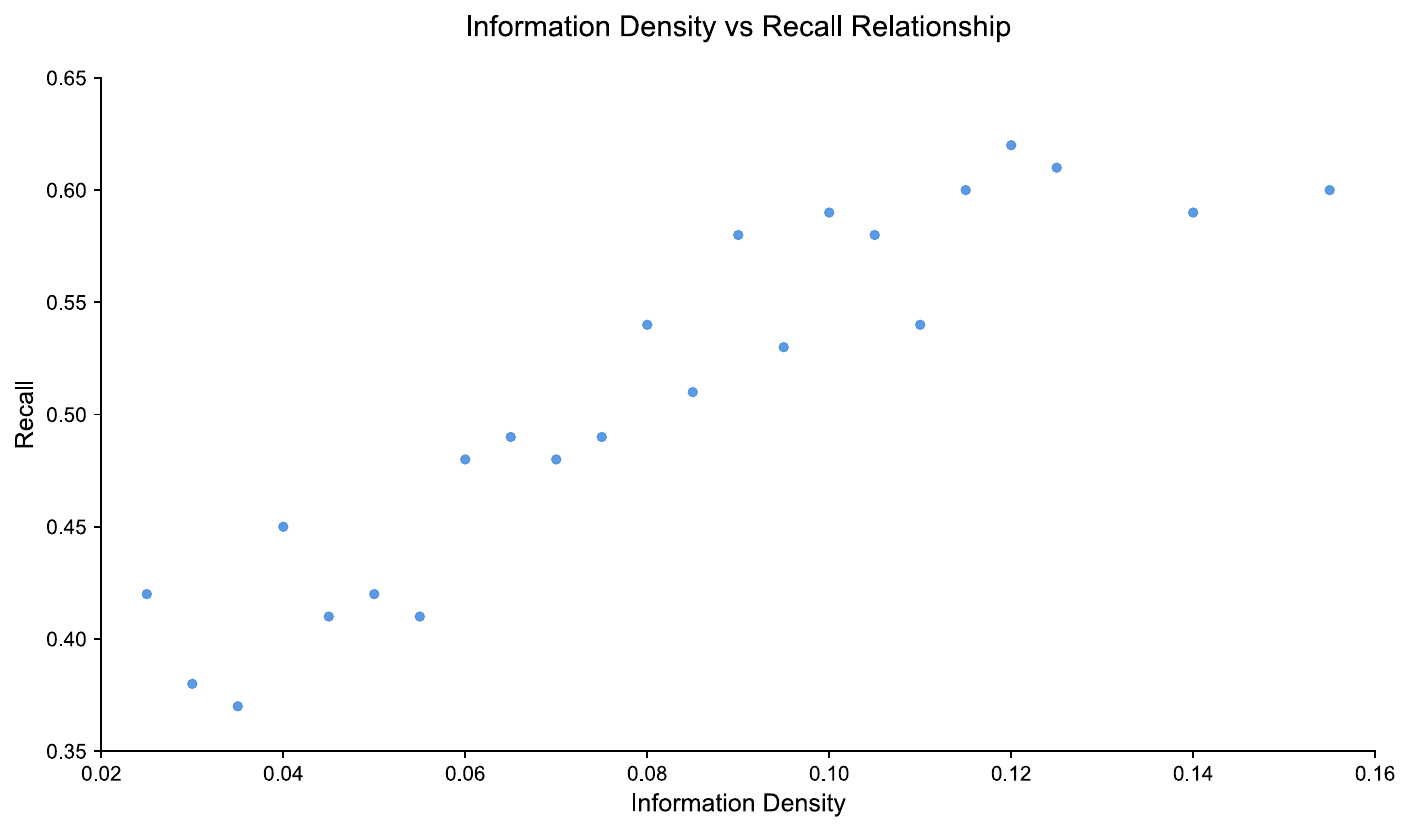}
\caption{Relationship between Information Density and Missed Detection Rate, $Recall$.}
\label{inforfig2}
\end{figure}

The large number of background tokens, as simple samples, dominates the model optimization process, leading to the following chain reaction: To minimize the overall loss, the gradient direction of the optimization algorithm is dominated by these massive background tokens, driving the model to prefer predicting more tokens as the 'O' label. This safe strategy maximizes the reduction of loss from incorrect entity predictions, causing the model's dependency on the 'O' label to intensify, but at the cost of sacrificing learning on rare entities. This statistical bias is not just simple data imbalance, but a specific manifestation of the long-tail problem in the NER task. By the training convergence phase, the model's generalization ability for low-frequency entity categories has been irreversibly affected. Even when facing entity samples with minor variations, the model tends to rely on its prior knowledge of high-frequency background tokens, resulting in a systematic misjudgment of entity words. This biased learning strategy directly results in a large number of False Negatives (FN), i.e., many entities that are misclassified as background tokens. According to the recall formula $Recall = TP / (TP + FN)$, a sharp increase in FN will inevitably cause a sharp decrease in Recall, thereby affecting performance. This is perfectly consistent with the results observed in the WNUT2017 subset experiments. Fig.\ref{inforfig2} strongly supports this mechanism, clearly showing that as information density decreases, the model's missed recognition rate significantly increases, and $Recall$ sharply drops. This pathway analysis profoundly reveals how background tokens in low-ID samples deeply affect model performance through training dynamics.

\subsubsection{The Attention Mechanism: A Weakening of Focus on Detail}
In the experiments, we observed that in the WNUT2017 dataset, 33.2\% of entities are multi-word combinations (e.g., “Empire State Building," “Clark Duke"), and the model struggles to accurately identify such entities in low-ID samples. The root cause of this phenomenon is the damage inflicted by low information density on the Transformer's core self-attention mechanism. The Transformer's self-attention mechanism allocates attention weights by calculating the similarity between Query and Key, with the core formula: $e_{ij} = (Q_i \cdot K_j^T) / \sqrt{d_k}$. Here, $Q_i$ represents the query vector of the current token, and $K_j$ represents the key vector of the $j$-th token in the sequence. In low-information-density text, the context often contains a large amount of redundant, bland background information. This causes the key vectors ($K_j$) generated for these semantically similar tokens to also become semantically homogeneous, lacking distinctiveness. Therefore, when a specific query vector $Q_i$ computes dot products with these highly similar $K_j$ vectors one by one, the resulting raw attention scores $e_{ij}$ will be very close. These similar scores, after Softmax normalization, produce a near-uniform attention probability distribution $a_{ij}$. We call this phenomenon \textbf{“attention blunting"}. It means the model cannot focus on the few truly important keywords or entity boundaries among the many background tokens; attention is smoothed and dispersed over the entire sentence. This severely weakens the model's ability to capture and integrate local key information, leading to poor performance on multi-word entity tasks that require precise boundary recognition, ultimately causing a drop in recall and F1-score.

To mathematically quantify and validate the “attention blunting" phenomenon, we borrowed from spectral analysis, which has been proven to be an effective tool for understanding attention mechanisms. We innovatively propose a new metric: \textbf{Attention Spectrum Analysis (ASA)}, with the specific definition and calculation process shown in Algorithm \ref{alg:asa}.

\begin{algorithm}[H]
\caption{Definition and Computation of the Attention-Spread Metric ASA}\label{alg:asa}
\begin{algorithmic}
\STATE \textbf{Input:}
\STATE \hspace{0.5em} $D={(x1,y1),\ (x2,y2),\ldots(xn,yn)}$ $\triangleright$ Dataset
\STATE \hspace{0.5em} $k$ $\triangleright$ frequency weight
\STATE \textbf{Output:}
\STATE \hspace{0.5em} Quantitative attention-distribution index ASA
\STATE 1: Load dataset D, fine-tune/train a BERT model, and obtain the attention matrix A.
\STATE 2: Apply the Fast Fourier Transform (FFT) to the attention matrix A, yielding FFT(A).
\STATE 3: Compute the power density spectrum of the frequency-domain representation, giving $|FFT(A)|^2$.
\STATE 4: To emphasize high-frequency components, multiply by the frequency weight $k$.
\STATE 5: Finally, the ASA index is calculated by summing the weighted power over all frequency components and normalizing:  
\begin{center} $ ASA = \frac{\sum_k \left( k \times |FFT(A)_k|^2 \right)}
           {\sum_k |FFT(A)_k|^2} $  \end{center}  
\end{algorithmic}
\end{algorithm}

From a signal processing perspective, the Transformer's attention weight distribution can be seen as an adaptive filter. An ideal attention distribution should be able to filter out key information. A sharp, focused attention distribution manifests as drastic, discontinuous changes in the weight sequence (e.g., a sudden increase in weight at entity boundaries), which corresponds to \textbf{high-frequency components} in the spectrum. A smooth, dispersed attention distribution manifests as gentle, continuous changes, corresponding to \textbf{low-frequency components}. High-frequency components correspond to highly discriminative information in the input sequence (e.g., entity word boundaries or key attributes), while low-frequency components correspond to stable, redundant background information. Therefore, the proportion of high-frequency components in the attention distribution's spectrum can effectively measure its ability to locate key information. A higher ASA value indicates a sharper attention distribution, a larger proportion of high-frequency components, and a stronger ability of the model to focus on and identify local details such as keywords and entity boundaries. A decrease in ASA directly reflects the degradation of the attention mechanism's performance as a filter, rendering it unable to effectively filter out noise. We use the above algorithm to calculate the ASA value, thereby validating that low information density alters the spectral properties of the attention distribution. As before, this section also uses stratified random sampling on the WNUT2017 dataset to construct subsets and statistically analyzes the relationship between the subset's information density and ASA. The experiment confirms that in low-ID samples, the ASA value is significantly lower. This directly verifies that low information density indeed causes attention distribution to become diffuse by weakening the high-frequency components in the attention spectrum, thus negatively impacting the model's ability to recognize multi-word entities and fine-grained semantic boundaries. In summary, the change in the ASA metric clearly reveals how information density affects the model's fine-grained recognition performance by influencing the attention distribution.

\begin{figure}[!htbp]
\centering
\includegraphics[width=\linewidth]{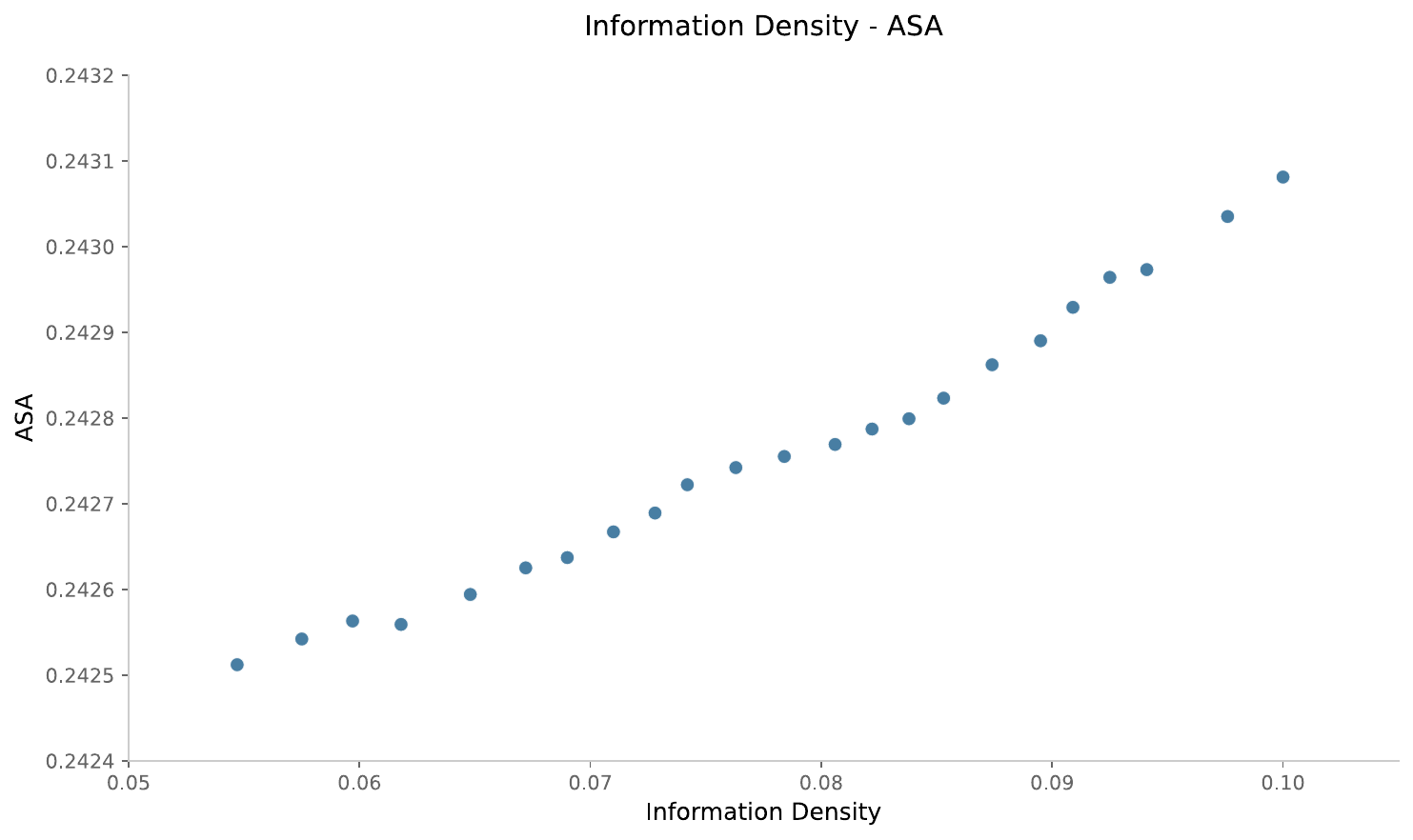}
\caption{Relationship between Information Density and ASA score.}
\label{asa}
\end{figure}

To validate the universality of the above mechanism, this experiment was extended to other typical UGC datasets, which share similar core features with WNUT2017: noisy text, sparse entities, all belonging to low-information-density scenarios. The mechanistic analysis experiments in the Appendix confirm that key bottleneck phenomena are reproduced in other UGC datasets: (1) \textbf{Recurrence of Statistical Conservative Bias:} Data statistics find that 'O' label background tokens also hold absolute dominance in these datasets, reproducing the extreme data imbalance that causes models to trend towards conservative predictions. (2) \textbf{Extension of Attention Blunting:} Calculation of their Attention Spectrum Analysis (ASA) values shows a distribution similar to WNUT2017, with ASA values generally lower in low-ID samples. This empirically confirms that the phenomenon of attention being diluted in homogeneous contexts also exists. This cross-dataset validation strongly demonstrates that “statistical conservative bias" and “attention blunting", driven by low information density, are not isolated phenomena of the WNUT2017 dataset but are common challenges for NER tasks in noisy UGC data.

Through an in-depth analysis of these two pathways, this section has systematically clarified the internal influence mechanism of information density on the performance of NER models for user-generated text. This analysis not only reveals the complete transmission chain of how low information density leads to a significant decline in model performance but also provides strong evidence that information density is the core issue at the heart of existing model performance bottlenecks.

\section{Model Architecture (WOM)}

The mechanistic analysis indicates that low information density is the core factor causing the performance bottleneck in NER for user-generated text. Its internal impact pathway is that such text regions induce “conservative prediction bias" and “attention blunting" in the model, thus systematically weakening its learning ability. To address this, this paper proposes the Window-aware Optimization Module (WOM), which fuses the windowing concept from long-text modeling and high-quality LLM-based back-translation technology. It uses a strategy combining regional detection and selective enhancement to directionally improve the effective information content of low-density regions, aiming to construct an information-richer learning environment for the model.

\begin{figure}[!htbp]
\centering
\centerline{\includegraphics[width=\linewidth]{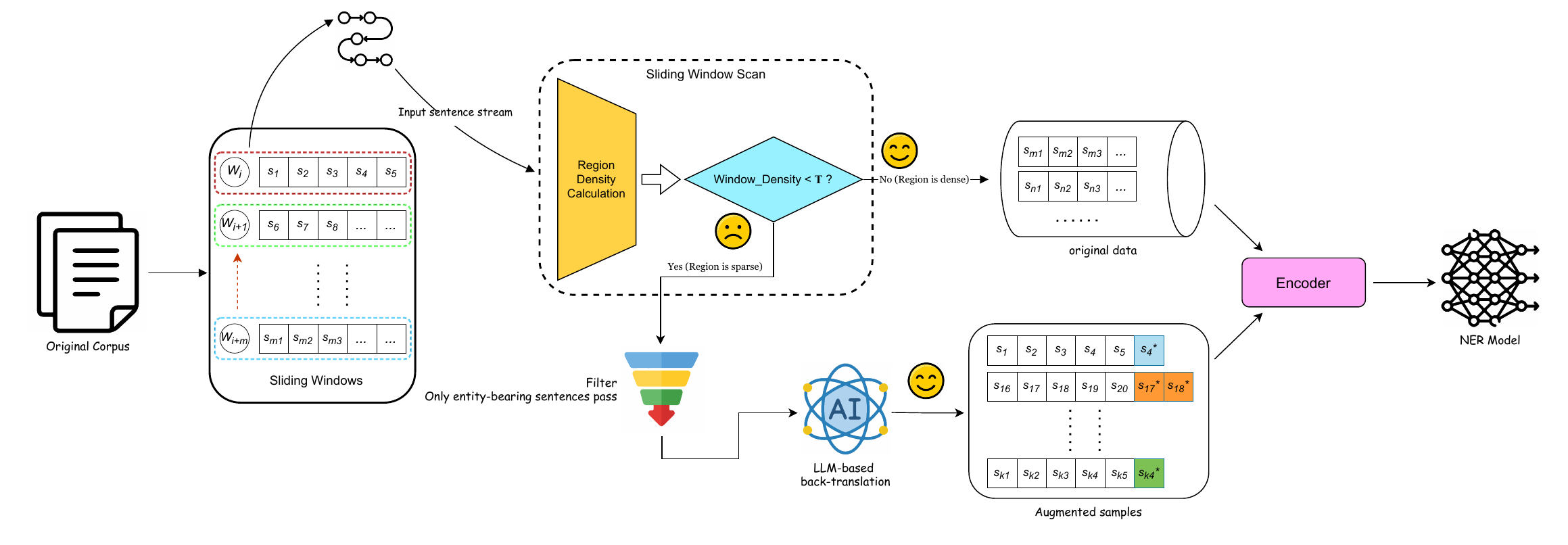}}
\caption{Module Operation Logic Diagram.}
\label{drawio}
\end{figure}

\subsection{Core Design Philosophy}
Targeting the mechanism by which low information density degrades model performance, the Window-aware Optimization Module (WOM) aims to intervene on the key variable—information density—at its source. Its core idea is to locate problematic regions via window scanning, and then directionally enhance the effective information density of those regions through selective back-translation, thereby blocking the transmission chain from low information density to model performance decline:

\begin{itemize}
    \item Regional Macro-Diagnosis: Identify low-density regions to locate the source of the problem. The negative impact of low information density is often concentrated in local text segments. Therefore, WOM first employs a sliding window to scan the entire training set, identifying information-barren regions based on an entity density threshold. Window-level diagnosis surpasses the limitations of single-sentence analysis and can capture contextual sparsity risks caused by continuous noisy text.
\end{itemize}
\begin{itemize}
    \item Targeted Micro-Intervention: Enhance entity signals to directionally improve effective information density. After locating problematic regions, WOM executes a selective enhancement strategy designed to address two core problems simultaneously: To counter the “statistical conservative bias" dominated by O-labels, we only augment sentences within the region that contain entities. This directly increases the relative proportion of entity samples in the local data distribution, breaking the model's predictive inertia and forcing it to pay attention to rare entity signals. To overcome “attention blunting" caused by homogeneous context, we use high-quality LLM-based back-translation technology. Compared to traditional translation tools, this can generate samples with consistent semantics, original entities, and labels, but with more diverse expressions (vocabulary, syntax). This artificially introduced information perturbation enriches the text's linguistic features, providing stronger guiding signals for the attention mechanism, allowing it to be reactivated and focus precisely even in a noisy background.WOM fuses a screening mechanism and LLM back-translation technology, and tests show it can effectively enhance the model's learning effect without requiring adjustments to the model structure, demonstrating good portability and generality.
\end{itemize}

\subsection{Module Architecture}
The Window-aware Optimization Module (WOM) is designed around the logic of precisely locating information-barren regions and directionally enhancing effective information density. WOM includes three core components (as illustrated in Fig.\ref{drawio}), covering the functions of data segmentation, problem region diagnosis, and density enhancement intervention, thereby effectively blocking the transmission path from low information density to performance degradation:

\subsubsection{Window Segmentation}
Used to decompose the training set text stream into segments for fine-grained evaluation. Following a predefined standard, we segment the dataset text stream into a series of non-overlapping windows of size W (sentence-level), ensuring comprehensive data coverage for the window detection engine.

\subsubsection{Window Detection Engine}
It is responsible for evaluating the information content of each window, calculating information density, and providing a quantitative score to ensure precise localization of information-barren text regions. If the window density $\le$ the information density threshold T, it is deemed information-barren and enters the enhancement phase; otherwise, it proceeds directly to the next window.

\subsubsection{Back-translation Enhancement Processing}
It employs back-translation technology to directionally enhance the effective information density of problem regions. For identified text-barren windows ($\le T$), WOM performs selective back-translation enhancement, applied \textit{only} to the sentences within it that carry entities. We first use an LLM to translate to an intermediate language (e.g., Chinese) and then back-translate to the original language, generating semantically consistent but syntactically diverse variants. To protect the core information, we apply an \textbf{entity-preservation mechanism}: entities are replaced with placeholders with their boundaries and categories locked before translation. After back-translation, we automatically verify if the entities are fully preserved and their labels are consistent; if not, the sample is discarded. Data that passes validation is concatenated with the original window to form new training samples. This process increases the window's $ET/TT$ while the $\Bigl(1 + \log \frac{TT}{TT_{SL}} \cdot \lambda\Bigr)$ term remains relatively stable, thus achieving a directional improvement in information density based on (\ref{deqn_ex1}).

\begin{figure}[!htbp]
\centering
\includegraphics[width=\linewidth]{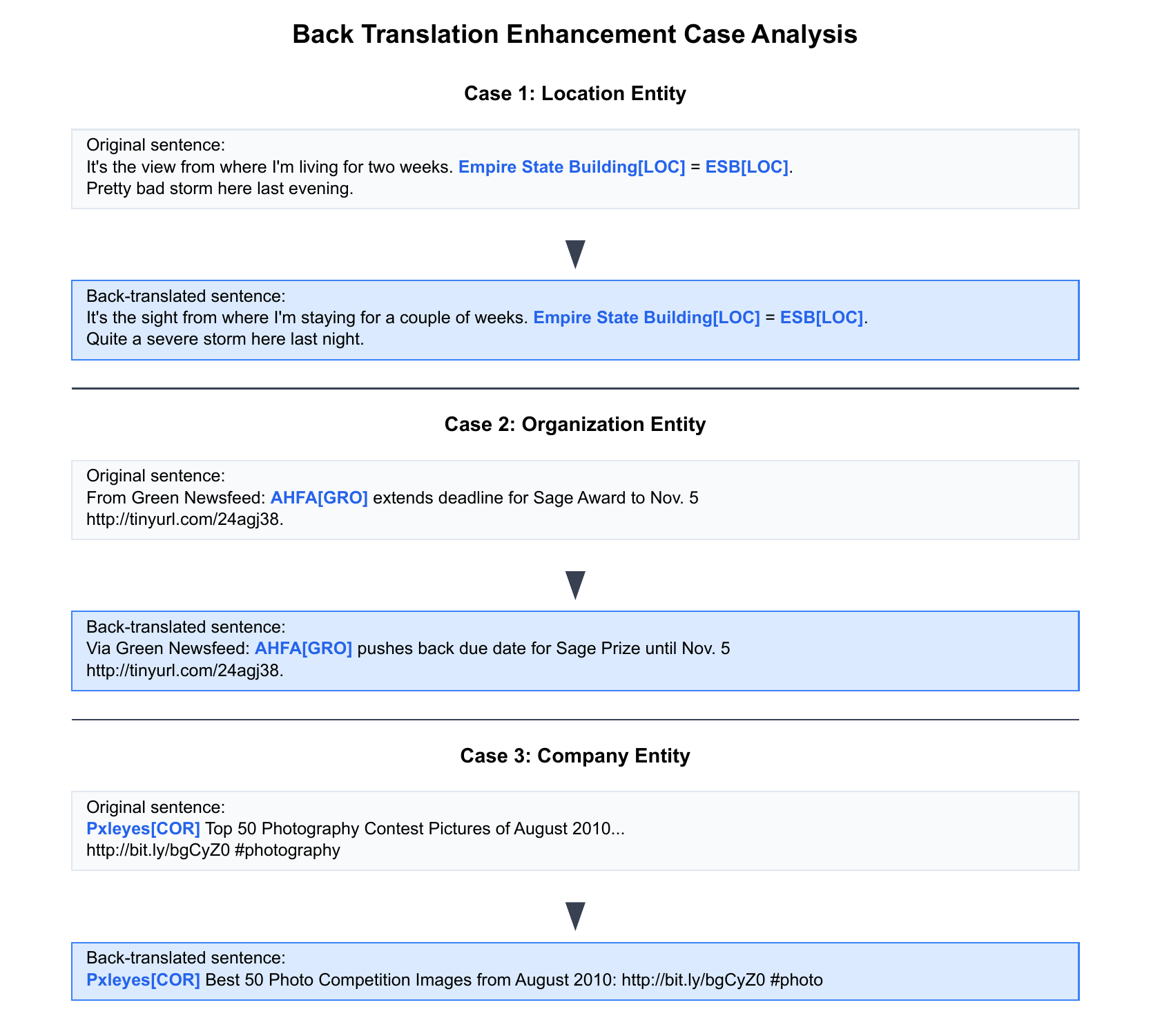}
\caption{Examples of LLM-based back-translated samples.}
\label{exampleBT}
\end{figure}

This mechanism ensures training data simultaneously incorporates both original and filtered back-translated samples, enabling efficient training. To visually demonstrate the module's operational mechanism, Fig.\ref{exampleBT} presents a comparison of an information-poor window before and after processing: Augmented samples strictly preserve core semantics and entity labels (e.g., “Empire State Building=ESB,” “AHFA,” etc.), while introducing reasonable variations in phrasing and vocabulary (e.g., replacing “living for two weeks” with “staying for a couple of weeks” and “extends deadline” with “pushes back due date”). This enhances local information density and expressive diversity.

\section{Experiment}
This section discusses in detail the effectiveness of WOM. Through a series of comparative experiments, we systematically analyze the module's performance advantages. In particular, this section conducts ablation studies and hyperparameter analysis to demonstrate the robustness and effectiveness of our method's design. The main performance evaluation, ablation studies, and hyperparameter analysis are conducted on WNUT2017, which is one of the most challenging benchmarks in the noisy text NER field. We also validate on two other widely used datasets of the same type, Twitter-Ner and WNUT2016. This section selects common pre-trained models for NER tasks as the basic framework, including the BERT and RoBERTa series, and builds and compares other models on this basis.

\subsection{Module Performance Evaluation}
To comprehensively evaluate the performance gains of WOM, this section applies it to multiple mainstream model architectures for exhaustive comparative experiments. Table \ref{tab:table2} shows the F1 results for each model. From this, it is evident that our proposed WOM, as an input-end optimization module, brings significant performance improvements when applied to several mainstream baseline models. For the core dataset WNUT2017 specifically: On the SpanNER (BERT-large) baseline, WOM brought a substantial 4.57\% improvement, increasing the F1-score from 51.2 to 55.77. On the RoBERTa (large)+BiLSTM backbone model, WOM increased the F1-score from 59.76 to 61.08, achieving a stable gain of 1.32\%. When combined with the more complex RoBERTa (large)+BiLSTM+CRF model, WOM still brought a 1.85\% performance improvement, with the F1-score reaching 61.39, achieving state-of-the-art results on this task at present. Furthermore, experiments on the Twitter-Ner and WNUT2016 datasets further validate WOM's generalization ability. As shown in Table \ref{tab:table2}, WOM also brought consistent performance improvements across different models. Synthesizing the results across multiple datasets, WOM, as a model-agnostic input enhancement module, effectively complements the inherent deficiencies of existing model architectures in handling information-sparse contexts by accurately diagnosing and intervening in low-information-density regions, thereby significantly enhancing model robustness in various noisy text environments.

\begin{table}[t]
\centering
\small
\footnotesize 
\sisetup{table-format = -1.3}
\begin{tabular}{cccc}
    \toprule
    \textbf{Model} & \textbf{WNUT2017} & \textbf{Twitter-Ner} & \textbf{WNUT2016} \\
    \midrule
    BERT\cite{devlin2019} & 48.14 & 72.61 & 48.49 \\
    \midrule
    RegLER(BERT)\cite{jeong2021regNER} & 52.74 & -- & -- \\
    \midrule
    SpanNER(BERT)\cite{fu2021spanner} & 51.2 & 74.93 & 53.17 \\
    \midrule
    MINER(BERT)\cite{wang2022miner} & 54.53 & 75.45 & 53.28 \\
    \midrule
    GoLLIE-7B\cite{sainz2024gollie} & 52.0 & -- & -- \\
    \midrule
    Context(RoBERTa)\cite{xu2023supplementary} & 59.03 & 76.7 & 56.55 \\
    \midrule
    CL-KL\cite{wang2021improving} & 60.45 & -- & 58.98 \\
    \midrule
    RoBERTa-BiLSTM-CRF\cite{RoBERTa-BiLSTM-CRF2024xu} & 59.54 & 77.85 & 59.03 \\
    \midrule
    SpanNER+\textbf{WOM} & \textbf{55.77} & \textbf{75.94} & \textbf{55.36} \\
    \midrule
    MINER+\textbf{WOM} & \textbf{55.88} & \textbf{76.24} & \textbf{56.67} \\
    \midrule
    Context+\textbf{WOM} & \textbf{60.91} & \textbf{77.47} & \textbf{60.69} \\
    \midrule
    RoBERTa-BiLSTM-CRF+\textbf{WOM} & \textbf{61.39} & \textbf{78.52} & \textbf{60.35} \\
    \bottomrule
  \end{tabular}
  \caption{A comparison of the performance of recent state-of-the-art models with that of our approach\label{tab:table2}}

  \vspace{0.5em} 
  \begin{minipage}{\linewidth}
    \small
    Note: The optimal choices of hyperparameters W and T vary slightly across different pre-trained models. In this work, we aim to select reasonably good values that deliver appreciable performance gains.
  \end{minipage}
\end{table}

\subsubsection{Ablation Study}
To validate the effectiveness of the Window-aware Optimization Module, this section designs an ablation study. The following ablation experiments use the Roberta(base)-BiLSTM model. There are three experimental settings: (1) Baseline: No data augmentation is used; the model is trained and evaluated directly on the original WNUT2017 training set. (2) Global Augmentation (GA): Back-translation enhancement is applied to the entire dataset text. This method does not use the window and detection engine and represents non-selective blind enhancement. This setting serves as a control group to verify the necessity of the selective enhancement strategy proposed in this paper. (3) Window-aware Optimization Module (WOM): The complete module proposed in this paper. After window segmentation, detection engine and threshold screening, targeted back-translation enhancement is performed on the barren window text.

\begin{itemize}
    \item Baseline: The model was trained and evaluated directly on the original WNUT2017 training set without any data augmentation.
\end{itemize}
\begin{itemize}
    \item Global Augmentation (GA): Back-translation was applied to the entire dataset. This approach, which omits the windowing and detection engine, represents a non-selective blind augmentation strategy. This setting was designed to contrast with our selective approach and demonstrate the necessity of a targeted and conditional strategy.
\end{itemize}
\begin{itemize}
    \item Window-oriented Optimization Module (WOM): Our complete proposed module was used, which involves partitioning the data into windows, using a detection engine with a threshold to identify information-scarce regions, and applying targeted back-translation only to those regions
\end{itemize}

\begin{table}[t]
  \centering
  \begin{tabular}{cccc}
    \toprule
    \textbf{Dataset} & \textbf{GA} & \textbf{WOM} & \textbf{F1-score} \\
    \midrule
    \multirow{3}{*}{WNUT2017} & $\times$ & $\times$ & 57.35 \\
    & $\checkmark$ & $\times$ & 56.35 \\
    & $\times$ & $\checkmark$ & \textbf{59.25} \\
    \midrule
    \multirow{3}{*}{Twitter-Ner} & $\times$ & $\times$ & 76.62 \\
    & $\checkmark$ & $\times$ & 76.22 \\
    & $\times$ & $\checkmark$ & \textbf{76.64} \\
    \midrule
    \multirow{3}{*}{WNUT2016} & $\times$ & $\times$ & 56.14 \\
    & $\checkmark$ & $\times$ & 56.02 \\
    & $\times$ & $\checkmark$ & \textbf{57.50} \\
    \bottomrule
  \end{tabular}
  \caption{Ablation study results on UGC datasets}\label{tab:method_performance}
\end{table}

The experimental results are shown in Table \ref{tab:method_performance}. For the WNUT2017 dataset, the baseline model's F1-score is 57.35. After applying global back-translation enhancement, the F1-score decreased to 56.35. The main reason for this performance drop is that the global enhancement strategy failed to distinguish the quality of the original text, introducing excessive uncontrollable noise and semantic perturbations. These unhealthy augmented samples not only failed to provide beneficial diversity but also had a negative impact on the model's learning process, disrupting the original data distribution. In contrast, our proposed complete WOM module significantly improved the model's performance to 59.25. This result stands in sharp contrast to the negative effect of global enhancement, clearly demonstrating the superiority of our proposed targeted, conditional enhancement strategy. Through our Window-aware Optimization Module, we can accurately identify those text segments that are information-barren and only perform targeted back-translation on these segments. This selective enhancement strategy successfully avoids the noise interference brought by global enhancement while effectively increasing information density. Furthermore, WOM also consistently outperforms the baseline and global enhancement strategies on the Twitter-Ner and WNUT2016 datasets, further proving the universality and superiority of our proposed density-aware optimization mechanism.

\subsubsection{Hyperparameter Analysis}
To validate WOM's robustness and to select reasonable hyperparameter configurations for the main experiments, this section provides a detailed analysis of its two core hyperparameters: the sliding window size $W$ and the information density threshold $T$. The following analysis is for the WNUT2017 dataset, using the Roberta(base)-BiLSTM model. It should be noted that the goal of this section is to determine a robustly good value for WOM under the current experimental setup, not a global optimum. In fact, due to differences in model architecture, the performance peak will fluctuate, indicating that there is no single universal optimal parameter for all models. \textbf{To ensure generalizability, the threshold $T$ is determined adaptively based on the average information density distribution of the training set. Similarly, the structure factor $\lambda$ serves as a smoothing factor derived from pilot experiments on a validation set.}

\begin{figure}[!htbp]
\centering
\includegraphics[width=3.2in]{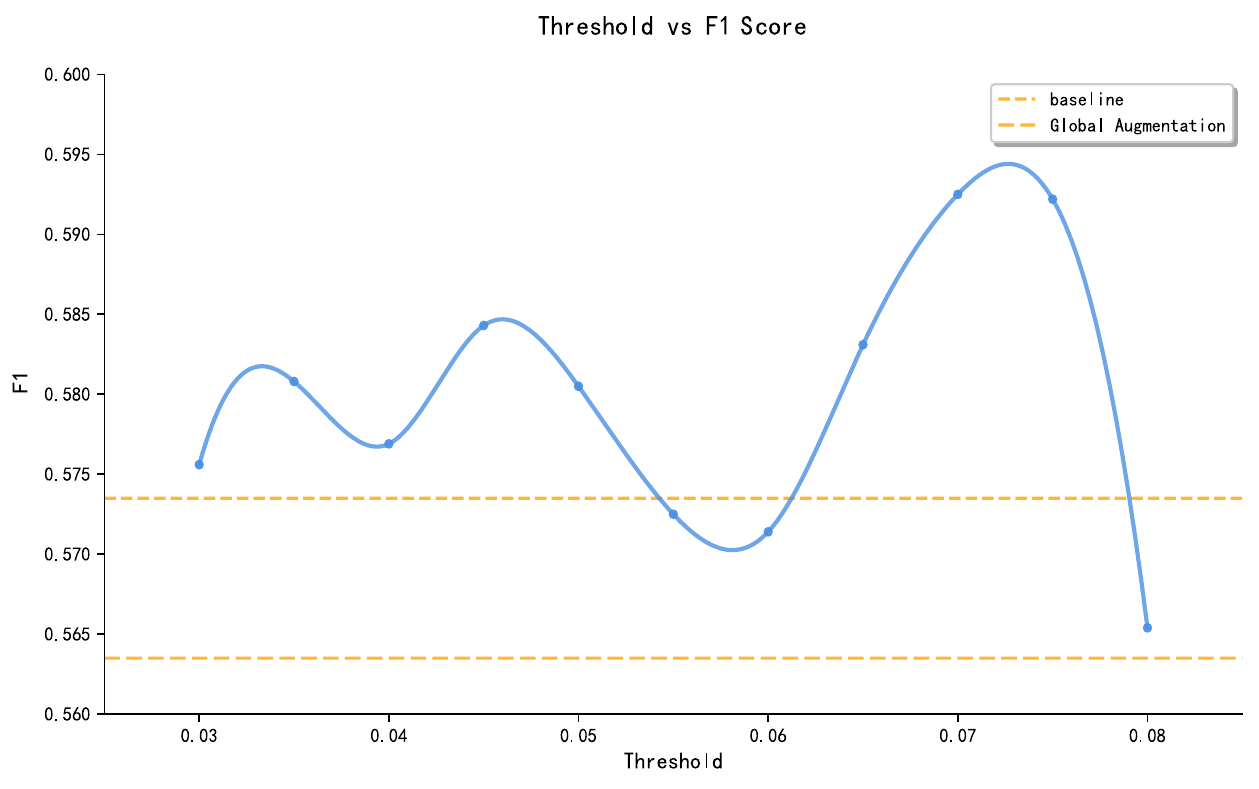}
\caption{This figure shows the F1 score performance under different thresholds (T).}
\label{yuhzi}
\end{figure}

\begin{itemize}
    \item Information Density Threshold (T): We initially set the window size to $W=30$ and adjust the threshold $T$ in the range [0.03, 0.08] with a step of 0.1, examining the effect of $T$ on model performance (F1-score). As shown in Fig.\ref{yuhzi}, at $T = 0.07$, WOM supplements the model with valuable, semantically diverse text, thereby significantly improving performance, and the model reaches a relatively optimal value. An excessively high $T$ value makes the trigger condition too loose, causing the model to perform unnecessary data augmentation on windows that already have acceptable information content, thereby introducing noise. Conversely, an excessively low $T$ value makes the trigger condition too strict, enhancing only windows with extremely low information density, leading to insufficient enhancement coverage. A large number of potential information-barren regions are not effectively augmented, so the improvement to model performance is limited, with results only slightly above the baseline.
\end{itemize}

\begin{figure}[!htbp]
\centering
\includegraphics[width=3in]{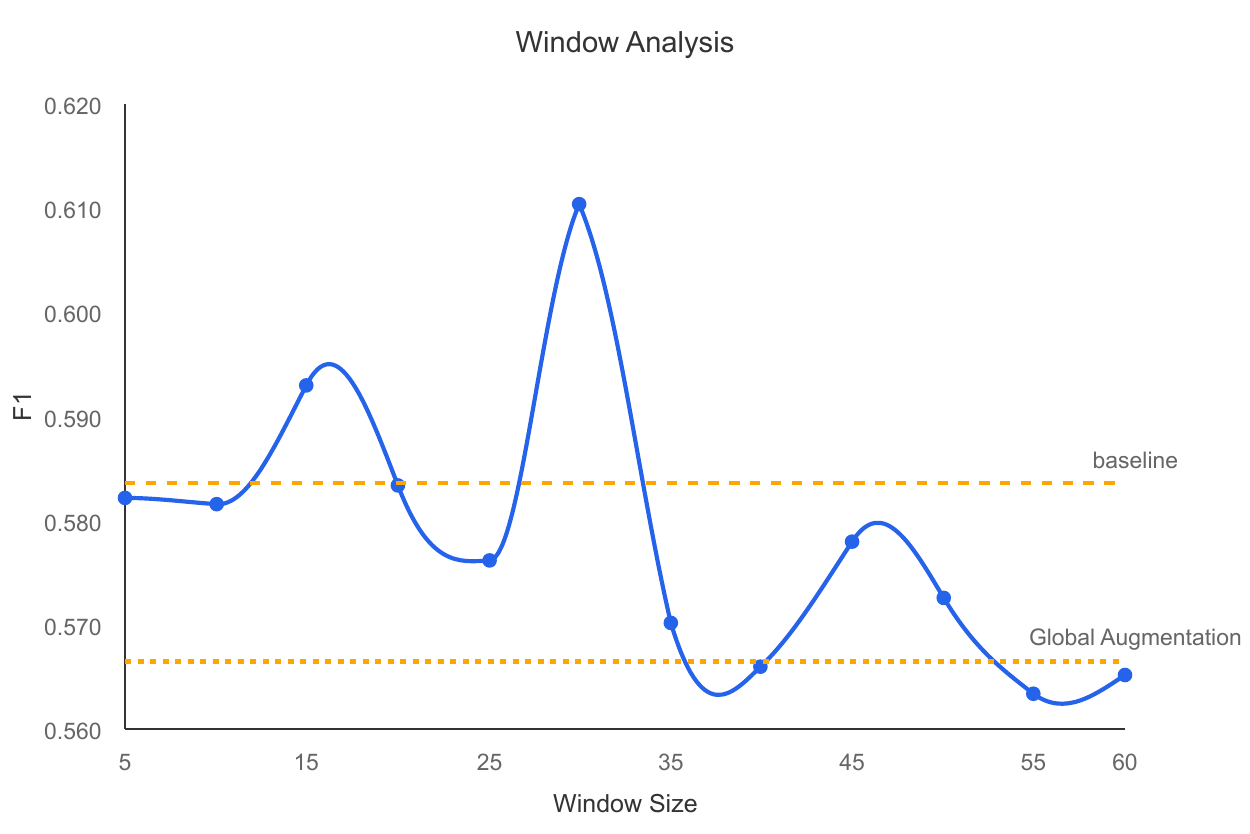}
\caption{This figure shows the F1 score performance under different window sizes (W).}
\label{window}
\end{figure}

\begin{itemize}
    \item Window Size (W): Similarly, as shown in Fig.\ref{window}, we fix the threshold $T=0.07$ and explore the impact of window size $W$ at different settings \{5, 10, ..., 60\}. It should be noted that a change in $W$ affects the calculation of local information density itself. The results show a similar pattern. An excessively small window (e.g., $W = 5$) provides an insufficient sample size for calculating information density. This causes the calculated information density to be overly sensitive to the characteristics of individual sentences, exhibiting high statistical instability. An excessively large window (e.g., $W \ge 50$) weakens the module's sensitivity to \textit{local} information-barren regions. Considering the stable effectiveness, we ultimately selected the relatively optimal point $W = 30$.
\end{itemize}

In summary, WOM is not extremely sensitive to its core hyperparameters—sliding window size $W$ and information density threshold $T$—but rather performs well within a reasonable range. Experiments demonstrate that an excessively small $W$ increases the variance of the information density estimation, while an excessively large $W$ weakens the detection sensitivity for local sparse segments; similarly, an overly high or low $T$ causes an imbalance between coverage and noise control. Overall, $W = 30$ and $T = 0.07$ are not only empirically effective but also theoretically justified, striking a good balance between the region's semantic richness and stability.

\section{Conclusion}
This paper systematically elucidates the critical role of Information Density (ID) in Named Entity Recognition (NER) for User-Generated Content (UGC), establishing a comprehensive research framework that spans conceptualization, mechanistic elucidation, and methodological optimization. Our findings not only establish a robust theoretical foundation for ``ID-driven text processing'' but also offer practical and deployable strategies for improving information extraction in complex and noisy scenarios.

The core contribution of this research lies in formalizing and validating ID as a pivotal structural feature that influences NER performance. Through correlation and sensitivity analyses, we demonstrate that ID significantly impacts the performance of various pre-trained models. We further introduce a frequency-domain diagnostic tool, \textbf{Attention Spectrum Analysis (ASA)}, to reveal the underlying mechanism of performance degradation: Low ID leads to ``attention decay'', which in turn results in compromised model efficacy. This provides a theoretical explanation for a critical performance bottleneck in NER.

Building upon this theoretical foundation, we design a \textbf{Window-Aware Optimization Module (WOM)} to suppress noise and enhance effective information density. Experiments demonstrate that WOM consistently achieves an absolute improvement of 1.0\%--4.5\% in F1-score across multiple UGC benchmarks and new state-of-the-art results on the WNUT2017 dataset. This validates the effectiveness of our ``mechanism-driven'' optimization strategy.

The academic value of this study extends beyond its methodological contributions. We advocate for the integration of ID as a standard attribute in UGC datasets and its standardized disclosure in experimental reports to foster research systematicity and reproducibility. Specifically, we recommend that future work report fine-grained results stratified by ID quantiles, in addition to overall metrics, to rigorously characterize differential model behaviors under varying information densities. This work also underscores the advantages of a ``mechanism-driven'' research paradigm, which, compared to traditional empirical tuning, is more targeted, interpretable, and efficacious. Our ID enhancement framework provides a scientific pathway for noise control, complementing other techniques such as data augmentation and domain adaptation.

For future work, we plan to deepen our research in the following four directions:

\begin{enumerate}
    \renewcommand{\labelenumi}{(\roman{enumi})} 
    
    \item \textbf{Learnable Causal ID Estimator:} Developing a learnable ID estimator based on causal inference to transcend the limitations of current proxy metrics, thereby enhancing the accuracy and reliability of information density assessment.
    
    \item \textbf{ASA Regularization Embedding:} Embedding the ASA method as a regularization term during the training process to proactively suppress attention decay in low-ID scenarios, further strengthening model robustness.
    
    \item \textbf{Dynamic WOM Optimization:} Exploring dynamic window strategies that integrate retrieval mechanisms, enabling WOM to achieve real-time adaptation to evolving data streams for online ID enhancement.
    
    \item \textbf{Extension to Other Information Extraction Tasks:} Extending the ID framework to related Information Extraction (IE) tasks, such as Relation Extraction and Entity Linking, to validate its generalizability and effectiveness across diverse IE domains.
\end{enumerate}

\appendix
\section{Supplement to Mechanism Analysis}
\label{app1}

\begin{figure}[!htbp]
\centering
\includegraphics[width=\linewidth]{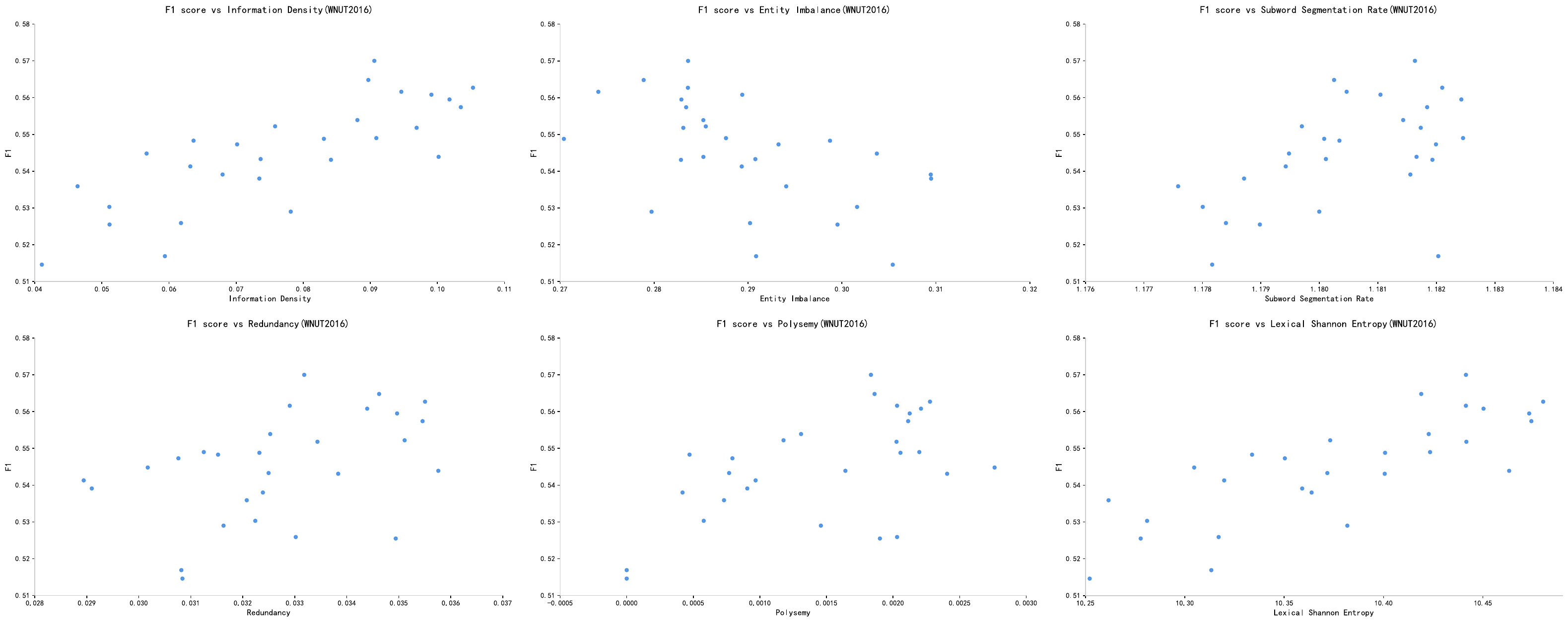}
\caption{Scatter plot of WNUT2016's structural features and performance.}
\label{fig12}
\end{figure}

We also conducted a systematic analysis and diagnosis of the six structural features proposed in this paper on the WNUT2016 dataset, another benchmark for noisy user-generated text (UGC), to validate the universality of our systematic diagnostic methodology. The Roberta-BiLSTM model was employed, with the process following the methodology outlined in the main text. The scatter plots for each structural feature are shown in Fig.\ref{fig12}, with correlation coefficients as follows.

\begin{figure}[!htbp]
\centering
\subfloat[]{\includegraphics[width=0.48\linewidth]{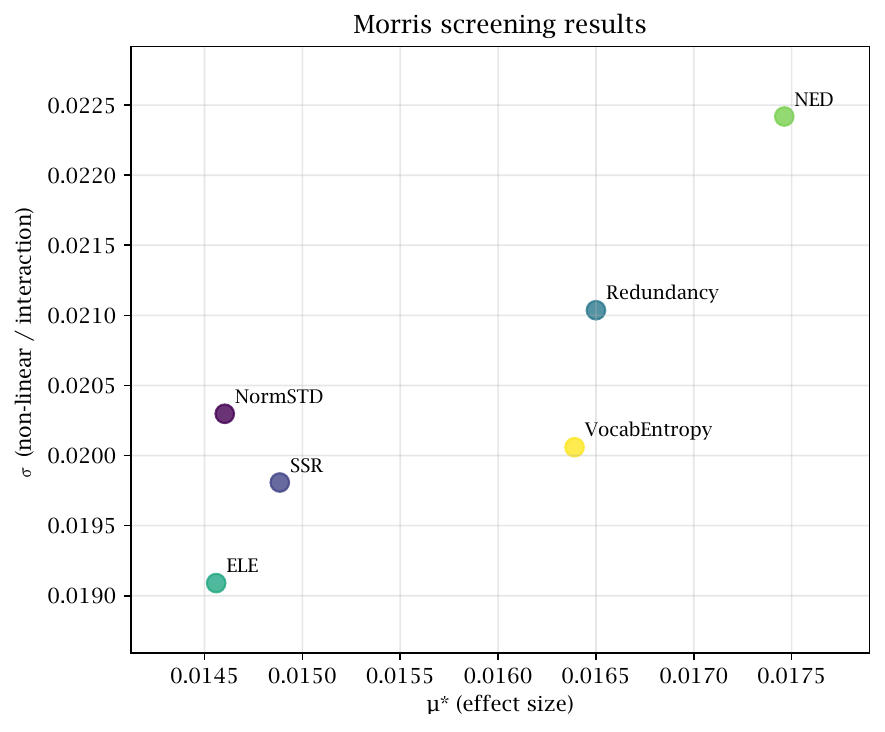}}
\label{fig_first_case3}
\hfil
\subfloat[]{\includegraphics[width=0.48\linewidth]{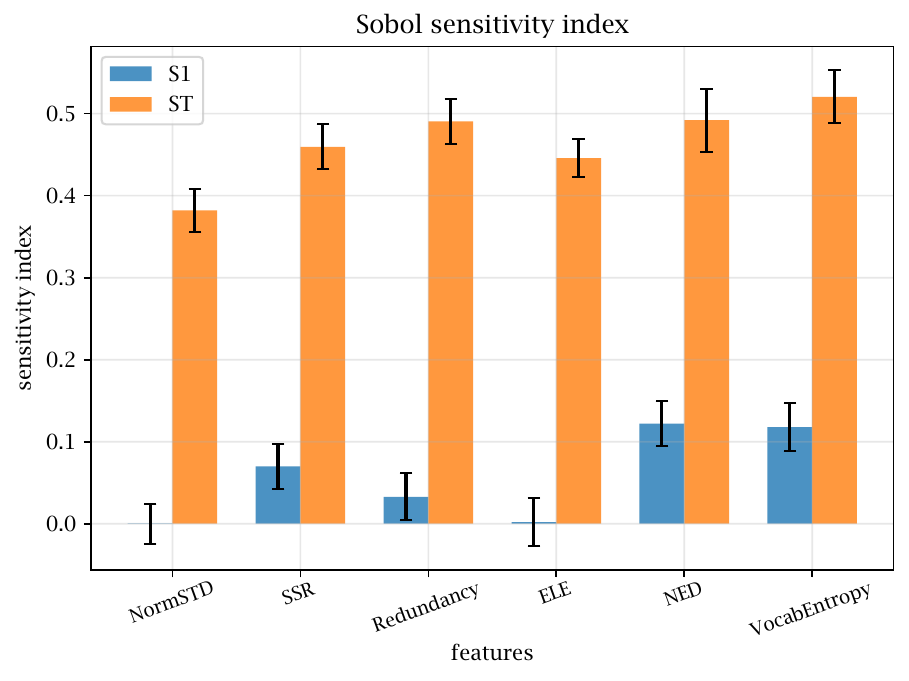}}
\label{fig_second_case3}
\caption{Sensitivity analysis results for different structural features of WNUT2016. (a) Morris. (b) Sobol.}
\label{fig13}
\end{figure}

Both the information density and lexical Shannon entropy correlation metrics exhibit significant results, with the former being a structural feature of the correlation index. The global sensitivity analysis results are shown in Fig.\ref{fig13}.

\begin{figure}[!htbp]
\centering
\includegraphics[width=\linewidth]{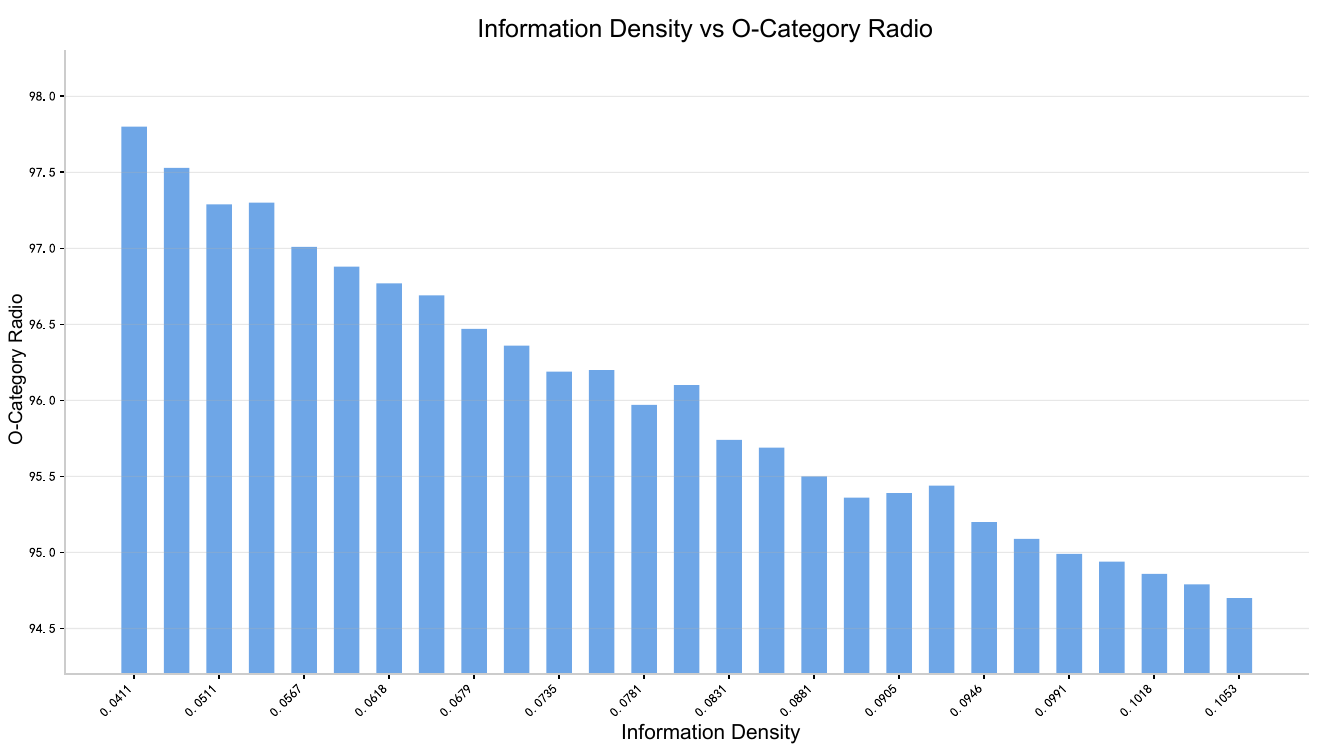}
\caption{Information Density vs. Proportion of O Category on WNUT2016.}
\label{fig14}
\end{figure}

\begin{figure}[!htbp]
\centering
\includegraphics[width=\linewidth]{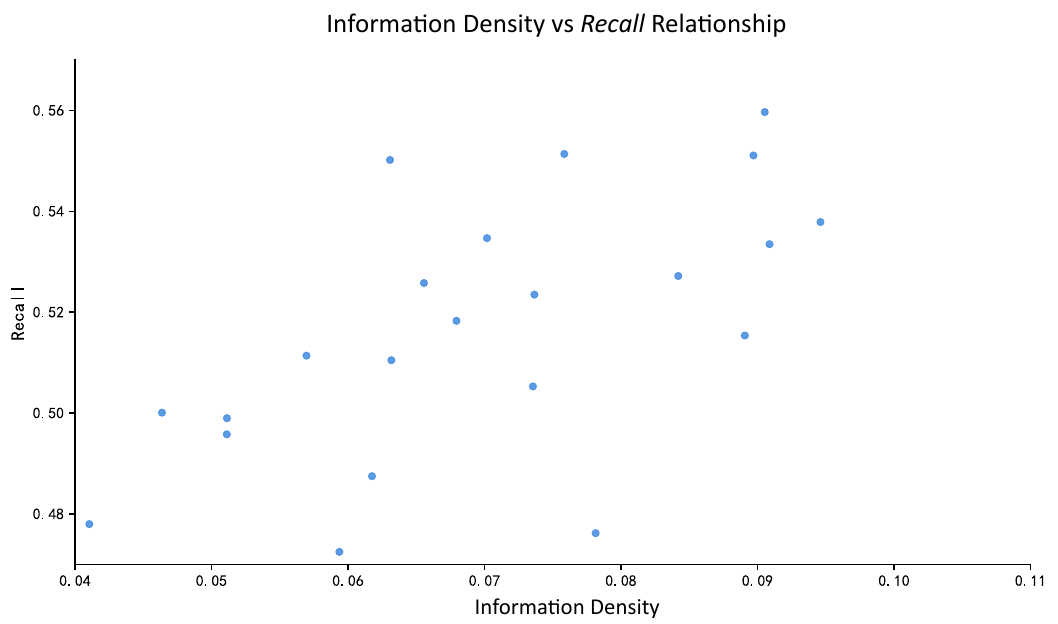}
\caption{Information Density vs. Recall on WNUT2016.}
\label{fig15}
\end{figure}

\begin{figure}[!htbp]
\centering
\includegraphics[width=\linewidth]{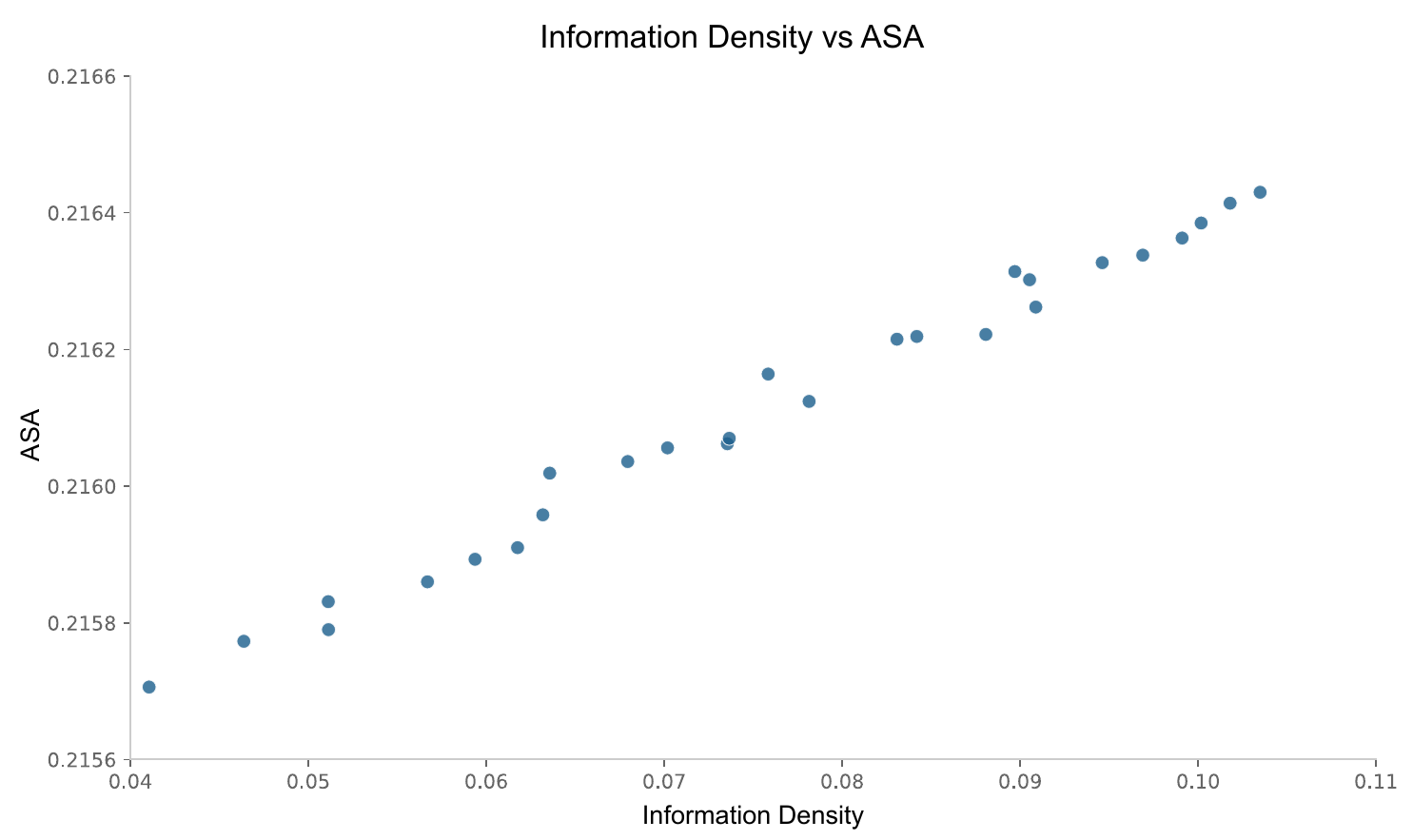}
\caption{ASA value distributions for different information densities on WNUT2016.}
\label{fig16}
\end{figure}

Similar to the analysis approach in the main text, the investigation of the impact mechanisms on WNUT2016 also follows two distinct pathways. Fig.\ref{fig14}, Fig.\ref{fig15}, and Fig.\ref{fig16} illustrate how low information density affects the loss function during model optimization and the operation of the internal self-attention mechanism within Transformers across different UGC datasets.
As illustrated, the trends depicted in the above images align with those presented in the main text. The mechanism whereby information density in the text affects model performance is equally applicable to other UGC datasets.

\begin{table}[!htbp]
\centering
\setlength{\tabcolsep}{4pt} 
\begin{tabular}{c c c}
    \hline
    \textbf{Structural Feature} & \textbf{Method} & \textbf{Coefficient} \\
    \hline
    \multirow{2}{*}{\centering Information Density} & Pearson & 0.812 \\
    & Sperman & 0.825 \\
    \hline
    \multirow{2}{*}{\centering Entity Imbalance Degree} & Pearson & -0.165 \\
    & Sperman & -0.144 \\
    \hline
    \multirow{2}{*}{\centering Subword Segmentation Rate} & Pearson & 0.28 \\
    & Sperman & 0.28 \\
    \hline
    \multirow{2}{*}{\centering Redundancy} & Pearson & 0.126 \\
    & Sperman & 0.111 \\
    \hline
    \multirow{2}{*}{\centering Entity Polysemy} & Pearson & 0.136 \\
    & Sperman & 0.081 \\
    \hline
    \multirow{2}{*}{\centering Lexical Shannon Entropy} & Pearson & 0.767 \\
    & Sperman & 0.785 \\
    \hline
\end{tabular}
\caption{Correlation Analysis of WNUT2016}\label{table4}
\end{table}

\bibliographystyle{elsarticle-num} 
\bibliography{citation.bib}


\end{document}